\documentclass[10pt,twocolumn,letterpaper]{article}

\newif\ifdouble
\doubletrue

\usepackage{cvpr}
\usepackage[utf8x]{inputenc}
\usepackage[T1]{fontenc}
\usepackage{ae,aecompl}
\usepackage[]{standalone}
\usepackage{tikz,pgfplots}
\usepackage{graphicx}
\usepackage{xcolor,colortbl}
\usepackage[caption=false]{subfig}
\usepackage{amsmath}
\usepackage{booktabs}
\usepackage{multirow}
\usepackage{diagbox}
\usepackage{balance}
\usepackage{amsthm}
\usepackage{amssymb}
\usepackage{amsfonts}
\usepackage{wasysym}
\usepackage{algorithm}
\usepackage{algorithmic}
\usepackage[breaklinks=true,colorlinks,bookmarks=false,pagebackref=true]{hyperref}

\usepackage{url}

\catcode`~=11

\catcode`~=13

\usepackage[numbers,sort&compress]{natbib}

\usetikzlibrary{arrows,automata,calc,shapes,shapes.symbols,positioning,fit,shadows}
\usetikzlibrary{pgfplots.statistics, pgfplots.colorbrewer} 
\usepgfplotslibrary{groupplots}

\pgfplotsset{
    compat=1.12,
    every legend to name picture/.style={scale=0.7},
    every legend to name picture/.style={font=\scriptsize},
}

\definecolor{mycolor1}{rgb}{0.85,0.85,1.0}
\definecolor{mycolor2}{rgb}{1.0,0.85,0.85}
\definecolor{mycolor3}{rgb}{0.0,0.75,0.75}%
\definecolor{mycolorg}{rgb}{0.5,1.00,0.5}
\definecolor{mycolorr}{rgb}{1.0,0.5,0.5}
\definecolor{mycolorm}{rgb}{1.0,0.0,1.0}%
\definecolor{green}{rgb}{0.0,0.5,0.0}
\definecolor{lightgreen}{rgb}{0.0,1.0,0.0}

\definecolor{color2}{rgb}{0.866666666666667,0.517647058823529,0.32156862745098}
\definecolor{color0}{rgb}{0.96,0.96,0.98}
\definecolor{color1}{rgb}{0.298039215686275,0.447058823529412,0.690196078431373}

\tikzset{
    >=stealth',
    line/.style = {draw, ->},
    narline/.style = {text width=2cm, font = \small},
    operation/.style={
           rectangle,
           rounded corners,
           draw=black,
           drop shadow={shadow scale=0.95},
           fill=green!10,
           minimum width=1.4cm,
           minimum height=1.1cm,
           text width=1.65cm,
           text centered},
    flow/.style={
           diamond,
           aspect=2,
           draw=black, thick,
           fill=blue!10,
           minimum height=1em,
           text centered},
    rect/.style={
           rectangle,
           draw=black, thick,
           minimum height=1em,
           text centered},
    input/.style={
           rectangle,
           draw=black, thick,
           minimum height=1em,
           text centered}
}

\title{Content Authentication for Neural Imaging Pipelines: End-to-end Optimization of Photo Provenance in Complex Distribution Channels}

\author{Paweł Korus$^{1,2}$ and Nasir Memon$^{1}$ \\
New York University$^{1}$, AGH University of Science and Technology$^{2}$\\
{\tt\small \url{http://kt.agh.edu.pl/~korus}}
}

\cvprfinalcopy

\def\cvprPaperID{4965} % *** Enter the CVPR Paper ID here

\begin{document}

\maketitle

\begin{abstract}
Forensic analysis of digital photo provenance relies on intrinsic traces left in the photograph at the time of its acquisition. Such analysis becomes unreliable after heavy post-processing, such as down-sampling and re-compression applied upon distribution in the Web. This paper explores end-to-end optimization of the entire image acquisition and distribution workflow to facilitate reliable forensic analysis at the end of the distribution channel. We demonstrate that neural imaging pipelines can be trained to replace the internals of digital cameras, and jointly optimized for high-fidelity photo development and reliable provenance analysis. In our experiments, the proposed approach increased image manipulation detection accuracy from 45\% to over 90\%. The findings encourage further research towards building more reliable imaging pipelines with explicit provenance-guaranteeing properties. 
\end{abstract}

\section{Introduction}
\label{sec:introduction}

\begin{figure*}[!t]
    \includegraphics[width=1.0\textwidth]{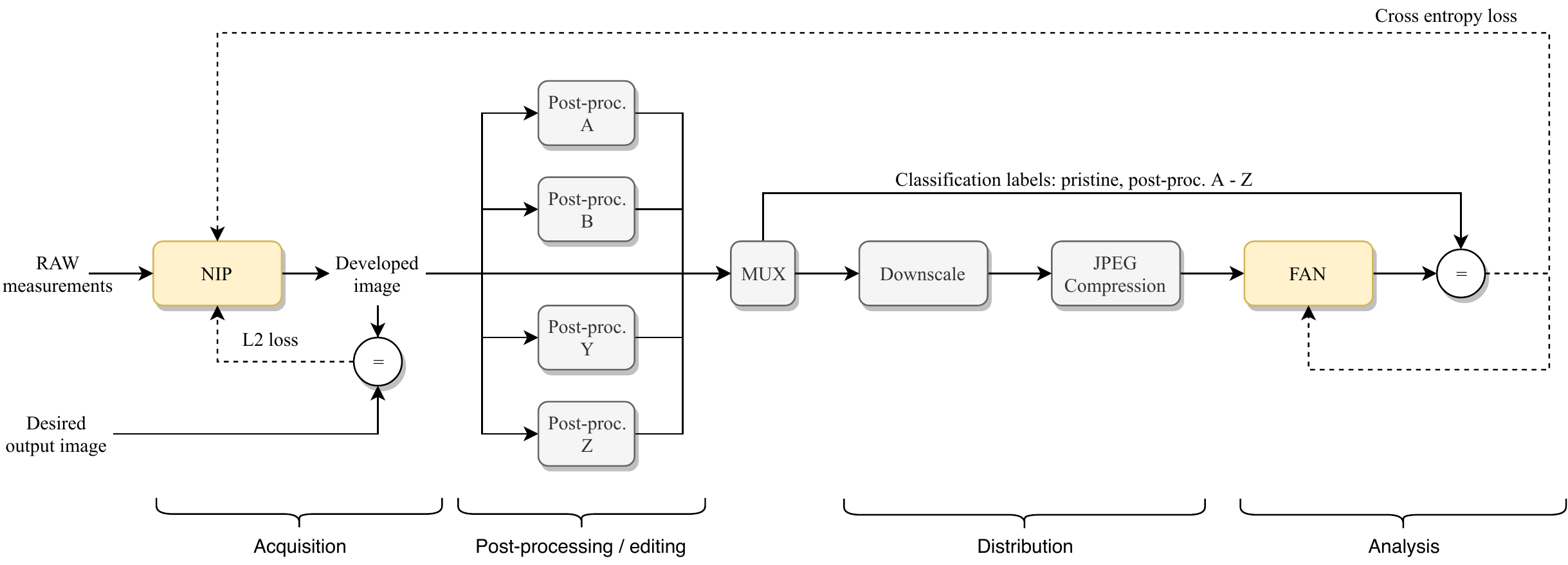}
    \caption{Optimization of the image acquisition and distribution channel to facilitate photo provenance analysis. The neural imaging pipeline (NIP) is trained to develop images that both resemble the desired target images, but also retain meaningful forensic clues at the end of complex distribution channels.}
    \label{fig:nip-training-protocol}
\end{figure*}

Ensuring integrity of digital images is one of the most challenging and important problems in multimedia communications. Photographs and videos are commonly used for documentation of important events, and as such, require efficient and reliable authentication protocols. Our current media acquisition and distribution workflows are built with entertainment in mind, and not only fail to provide explicit security features, but actually work against them. Image compression standards exploit heavy redundancy of visual signals to reduce communication payload, but optimize for human perception alone. Security extensions of popular standards lack in adoption~\cite{jpegSecurity}.

Two general approaches to assurance and verification of digital image integrity include~\cite{korus2017digital,piva2013overview,stamm2013information}: (1) pro-active protection methods based on digital signatures or watermarking; (2) passive forensic analysis which exploits inherent statistical properties resulting from the photo acquisition pipeline. While the former provides superior performance and allows for advanced protection features (like precise tampering localization~\cite{yan2017multi}, or reconstruction of tampered content~\cite{Korus2014b}), it failed to gain widespread adoption due to the necessity to generate protected versions of the photographs, and the lack of incentives for camera vendors to modify camera design to integrate such features~\cite{blythe2004secure}.

Passive forensics, on the other hand, relies on our knowledge of the photo acquisition pipeline, and statistical artifacts introduced by its successive steps. While this approach is well suited for potentially analyzing any digital photograph, it often falls short due to the complex nature of image post-processing and distribution channels. Digital images are not only heavily compressed, but also enhanced or even manipulated before, during or after dissemination. Popular images have many online incarnations, and tracing their distribution and evolution has spawned a new field of image phylogeny~\cite{dias2013large,dias2013toward} which relies on visual differences between multiple images to infer their relationships and editing history. However, phylogeny does not provide any tools to reason about the authenticity or history of individual images. Hence, reliable authentication of real-world online images remains untractable~\cite{zampoglou2016large}. 

At the moment, forensic analysis often yields useful results in near-acquisition scenarios. Analysis of native images straight from the camera is more reliable, and even seemingly benign implementation details - like the rounding operators used in the camera's image signal processor~\cite{agarwal2017photo} - can provide useful clues. Most forensic traces quickly become unreliable as the image undergoes further post-processing. One of the most reliable tools at our disposal involves analysis of the imaging sensor's artifacts (the photo response non-uniformity pattern) which can be used both for source attribution and content authentication problems~\cite{Chen2008}.

In the near future, rapid progress in computational imaging will challenge digital image forensics even in near-acquisition authentication. In the pursuit of better image quality and convenience, digital cameras and smartphones employ sophisticated post-processing directly in the camera, and soon few photographs will resemble the original images captured by the sensor(s). Adoption of machine learning has recently challenged many long-standing limitations of digital photography, including: (1) high-quality low-light photography~\cite{Chen2018}; (2) single-shot HDR with overexposed content recovery~\cite{Eilertsen2017}; (3) practical high-quality digital zoom from multiple shots~\cite{super-resolution-pixel3}; (4) quality enhancement of smartphone-captured images with weak supervision from DSLR photos~\cite{ignatov2017wespe}.

These remarkable results demonstrate tangible benefits of replacing the entire acquisition pipeline with neural networks. As a result, it will be necessary to investigate the impact of the emerging \emph{neural imaging pipelines} on existing forensics protocols. While important, such evaluation can be seen rather as damage assessment and control and not as a solution for the future. We believe it is imperative to consider novel possibilities for security-oriented design of our cameras and multimedia dissemination channels. 

In this paper, we propose to optimize neural imaging pipelines to improve photo provenance in complex distribution channels. We exploit end-to-end optimization of the entire photo acquisition and distribution channel to ensure that reliable forensics decisions can be made even after complex post-processing, where classical forensics fails (Fig.~\ref{fig:nip-training-protocol}). We believe that imminent revolution in camera design creates a unique opportunity to address some of the long-standing limitations of the current technology. While adoption of digital watermarking in image authentication was limited by the necessity to modify camera hardware, our approach exploits the flexibility of neural networks to learn relevant integrity-preserving features within the expressive power of the model. We believe that with solid security-oriented understanding of neural imaging pipelines, and with the rare opportunity of replacing the well-established and security-oblivious pipeline, we can significantly improve digital image authentication capabilities. 

We aim to inspire discussion about novel camera designs that could improve photo provenance analysis capabilities. We demonstrate that it is possible to optimize an imaging pipeline to significantly improve detection of photo manipulation at the end of a complex real-world distribution channel, where state-of-the-art deep-learning techniques fail. The main contributions of our work include:
\begin{enumerate}
    \itemsep0em
    \item The first end-to-end optimization of the imaging pipeline with explicit photo provenance objectives;
    \item The first security-oriented discussion of neural imaging pipelines and the inherent trade-offs;
    \item Significant improvement of forensic analysis performance in challenging, heavily post-processed conditions;
    \item A neural model of the entire photo acquisition and distribution channel with a fully differentiable approximation of the JPEG codec. 
\end{enumerate}

To facilitate further research in this direction, and enable reproduction of our results, our neural imaging toolbox is available at \url{https://github.com/pkorus/neural-imaging}.

\section{Related Work}
\label{sec:related-work}

\paragraph{Trends in Pipeline Design} Learning individual steps of the imaging pipeline (e.g., demosaicing) has a long history~\cite{kapah2000demosaicking} but regained momentum in the recent years thanks to adoption of deep learning. Naturally, the research focused on the most difficult operations, i.e., demosaicing~\cite{Gharbi2016,Tan2017,Kokkinos2018,Syu2018} and denoising~\cite{burger2012image,zhang2017beyond,lehtinen2018noise2noise}. Newly developed techniques delivered not only improved performance, but also additional features. Gharbi et al. proposed a convolutional neural network (CNN) trained for joint demosaicing and denoising~\cite{Gharbi2016}. A recent work by Syu et al. proposes to exploit CNNs for joint optimization of the color filter array and a corresponding demosaicing filter~\cite{Syu2018}. 

Optimization of digital camera design can go even further. The recently proposed L3 model by Jiang et al. replaces the entire imaging pipeline with a large collection of local linear filters~\cite{Jiang2017}. The L3 model reproduces the entire photo development process, and aims to facilitate research and development efforts for non-standard camera designs. In the original paper, the model was used for learning imaging pipelines for RGBW (red-green-blue-white) and RGB-NIR (red-green-blue-near-infra-red) color filter arrays. 

Replacing the entire imaging pipeline with a modern CNN can also overcome long-standing limitations of digital photography. Chen et al. trained a UNet model~\cite{ronneberger2015u} to develop high-quality photographs in low-light conditions~\cite{Chen2018} by exposing it to paired examples of images taken with short and long exposure. The network learned to develop high-quality well-exposed color photographs from underexposed raw input, and yielded  better performance than traditional image post-processing based on brightness adjustment and denoising. Eilertsen et al. also trained a UNet model to develop high-dynamic range images from a single shot~\cite{Eilertsen2017}. The network not only learned to correctly perform tone mapping, but was also able to recover overexposed highlights. This significantly simplifies HDR photography by eliminating the need for bracketing and dealing with ghosting artifacts.  

\vspace{-12pt}
\paragraph{Trends in Forensics} The current research in forensic image analysis focuses on two main directions: (1) learning deep features relevant to low-level forensic analysis for problems like manipulation detection~\cite{bayar2018constrained,zhou2018}, identification of the social network of origin~\cite{amerini2017tracing}, camera model identification~\cite{cozzolino2018noiseprint}, or detection of artificially generated content~\cite{li2018detection}; (2) adoption of high-level vision to automate manual analysis that exposes physical inconsistencies, such as reflections~\cite{sun2017object,wengrowski2017reflection}, or shadows~\cite{kee2014exposing}. To the best of our knowledge, there are currently no efforts to either assess the consequences of the emerging neural imaging pipelines, or to exploit this opportunity to improve photo reliability.

\section{End-to-end Optimization of Photo Provenance Analysis}
\label{sec:pipeline-optimization}

\begin{figure*}[!t]
    \centering
    \includegraphics[width=1.00\textwidth]{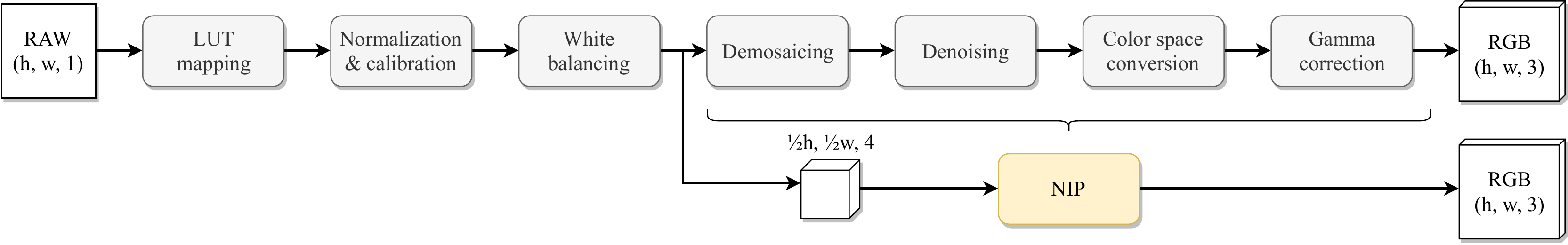}
    \vspace{0.0cm}
    \caption{Adoption of a neural imaging pipeline to develop raw sensor measurements into color RGB images: (top) the standard imaging pipeline; (bottom) adoption of the NIP model.}
    \label{fig:pipeline}
\end{figure*}

Digital image forensics relies on intrinsic statistical artifacts introduced to photographs at the time of their acquisition. Such traces are later used for reasoning about the source, authenticity and processing history of individual photographs. The main problem is that contemporary media distribution channels employ heavy compression and post-processing which destroy the traces and inhibit forensic analysis.

The core of the proposed approach is to model the entire acquisition and distribution channel, and optimize the neural imaging pipeline (NIP) to facilitate photo provenance analysis after content distribution (Fig.~\ref{fig:nip-training-protocol}). The analysis is performed by a forensic analysis network (FAN) which makes a decision about the authenticity/processing history of the analyzed photograph. In the presented example, the model is trained to perform manipulation detection, i.e., aims to classify input images as either coming straight from the camera, or as being affected by a certain class of post-processing. The distribution channel mimics the behavior of modern photo sharing services and social networks which habitually down-sample and re-compress the photographs. As will be demonstrated later, forensic analysis in such conditions is severely inhibited.

The parameters of the NIP are updated to guarantee both faithful representation of a desired color photograph ($L_2$ loss), and accurate decisions of forensics analysis at the end of the distribution channel (cross-entropy loss). Hence, the parameters of the NIP and FAN models are chosen as:
\begin{subequations}
    \begin{align}
	\theta_{\text{nip}}^{*} &= \underset{\theta_{\text{nip}}}{\operatorname{argmin}}~ \sum_{n} \Big( \| y_n - \text{nip}(x_n~|~\theta_{\text{nip}}) \|_2 \\ &+ \sum_{c} \text{log} \big( \text{fan}_{c} (d_c( \text{nip}(x_n~|~\theta_{\text{nip}})  )~|~\theta_{\text{fan}}) \big) \Big)
\end{align}
\end{subequations}
\begin{equation*}
	\theta_{\text{fan}}^{*} = \underset{\theta_{\text{fan}}}{\operatorname{argmin}}~ \sum_{n} \sum_{c} \text{log} \big( \text{fan}_{c} (d_c( \text{nip}(x_n~|~\theta_{\text{nip}})  )~|~\theta_{\text{fan}}) \big) 
\end{equation*}
\noindent where: $\theta_{\text{nip/fan}}$ are the parameters of the NIP and FAN networks, respectively; $x_n$ are the raw sensor measurements for the $n$-th example patch; $y_n$ is the corresponding target color image; $\text{nip}(x_n)$ is the color RGB image developed by NIP from $x_n$; $d_c()$ denotes a color image patch processed by manipulation $c$; $\text{fan}_c()$ is the probability that an image belongs to the $c$-th manipulation class, as estimated by the FAN model.

\subsection{The Neural Imaging Pipeline}

We replace the entire imaging pipeline with a CNN which develops raw sensor measurements into color RGB images (Fig.~\ref{fig:pipeline}). Before feeding the images to the network, we pre-process them by reversing the nonlinear value mapping according to the camera's LUT, subtracting black levels from the edge of the sensor, normalizing by sensor saturation values, and applying white-balancing according to shot settings. We also standardized the inputs by reshaping the tensors to have feature maps with successive measured color channels. This ensures a well-formed input of shape ($\frac{h}{2}, \frac{w}{2}, 4$) with values normalized to [0, 1]. All of the remaining steps of the pipeline are replaced by a NIP. See Section~\ref{sec:pipelines} for details on the considered pipelines. 

\begin{figure*}[!t]
    \includegraphics[width=1.0\textwidth]{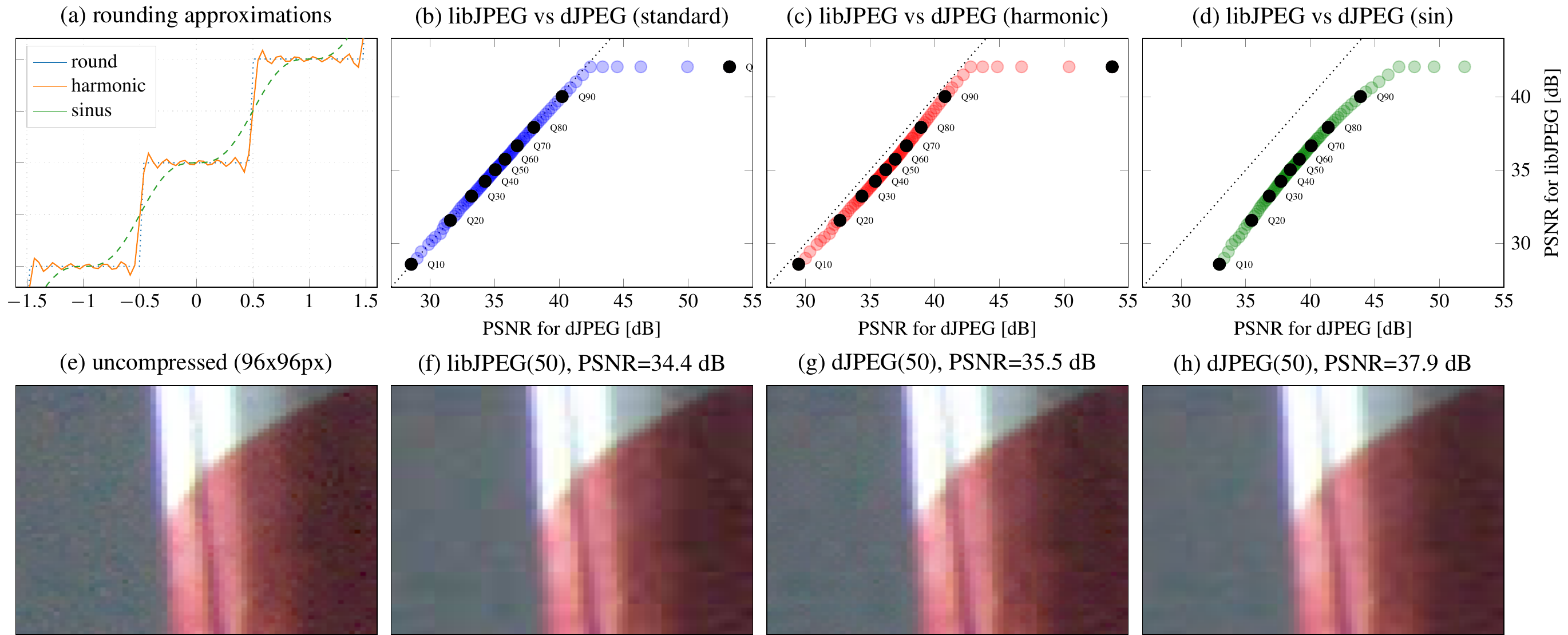}
    \vspace{0.0cm}
    \caption{Implementation of JPEG compression as a fully differentiable \emph{dJPEG} module: (a) continuous approximations of the rounding function; (b)-(d) validation of the dJPEG module against the standard \emph{libJPEG} library with standard rounding, and the harmonic and sinusoidal approximations; (e) an example image patch; (f) standard JPEG compression with quality 50; (g)-(h) dJPEG-compressed patches with the harmonic and sinusoidal approximations.}
    \label{fig:jpegnet}
\end{figure*}

\subsection{Approximation of JPEG Compression}

To enable end-to-end optimization of the entire acquisition and distribution channel, we need to ensure that every processing step remains differentiable. In the considered scenario, the main problem is JPEG compression. We designed a \emph{dJPEG} model which approximates the standard codec, and expresses its successive steps as matrix multiplications or convolution layers that can be implemented in TensorFlow (see supplementary materials for a detailed network definition):
\begin{itemize}
    \setlength\itemsep{0em}
    \item RGB to/from YCbCr color-space conversions are implemented as $1\times1$ convolutions.
    \item Isolation of $8\times8$ blocks for independent processing is implemented by a combination of \emph{space-to-depth} and reshaping operations.
    \item Forward/backward 2D discrete cosine transforms are implemented by matrix multiplication according to $DxD^{T}$ where $x$ denotes a $8\times8$ input, and $D$ denotes the transformation matrix.
    \item Division/multiplication of DCT coefficients by the corresponding quantization steps are implemented as element-wise operations with properly tiled and concatenated quantization matrices (for both the luminance and chrominance channels).
    \item The actual quantization is approximated by a continuous function $\rho(x)$(see details below).
\end{itemize}

The key problem in making JPEG differentiable lies in the rounding of DCT coefficients. We considered two approximations (Fig.~\ref{fig:jpegnet}a). Initially, we used a Taylor series expansion, which can be made arbitrarily accurate by including more terms. Finally, we used a smoother, and simpler sinusoidal approximation obtained by matching its phase with the sawtooth function:
\begin{equation}
    \rho(x) = x -  \frac{\text{sin}(2 \pi x)}{2\pi}
\end{equation}

We validated our \emph{dJPEG} model by comparing produced images with a reference codec from \emph{libJPEG}. The results are shown in Fig.~\ref{fig:jpegnet}bcd for a standard rounding operation, and the two approximations, respectively. We used 5 terms for the harmonic rounding. The developed module produces equivalent compression results with standard rounding, and a good approximation for its differentiable variants. Fig.~\ref{fig:jpegnet}e-h show a visual comparison of an example image patch, and its \emph{libJPEG} and \emph{dJPEG}-compressed counterparts.

In our distribution channel, we used quality level 50.

\subsection{The Forensic Analysis Network}

The forensic analysis network (FAN) is implemented as a CNN following the most recent recommendations on construction of neural networks for forensics analysis~\cite{bayar2018constrained}. Bayar and Stamm proposed a new layer type, which constrains the learned filters to be valid residual filters~\cite{bayar2018constrained}. Adoption of the layer helps ignore visual content and facilitates extraction of forensically-relevant low-level features. In summary, our network operates on $128 \times 128 \times 3$ patches in the RGB color space and includes (see supplement for full network definition):
\begin{itemize}
    \itemsep0em 
    \item A constrained convolutions layer learning $5 \times 5$ residual filters and with no activation function.
    \item Four $5 \times 5$ convolutional layers with doubling number of output feature maps (starting from 32). The layers use leaky ReLU activation and are followed by $2\times2$ max pooling.
    \item A $1\times1$ convolutional layer mapping 256 features into 256 features.
    \item A global average pooling layer reducing the number of features to 256.
    \item Two fully connected layers with 512 and 128 nodes activated by leaky ReLU.
    \item A fully connected layer with $N=5$ output nodes and softmax activation.
\end{itemize}

\noindent The network has 1,341,990 parameters in total, and outputs probabilities of 5 possible processing histories (4 manipulation classes \& straight from camera).

\section{Experimental Evaluation}

We started our evaluation by using several NIP models to reproduce the output of a standard imaging pipeline (Sec.~\ref{sec:pipelines}). Then, we used the FAN model to detect popular image manipulations (Sec.~\ref{sec:manipulation}). Initially, we validated that the models work correctly by using it without a distribution channel (Sec.~\ref{sec:fan-validation}). Finally, we performed extensive evaluation of the entire acquisition and distribution network (Sec.~\ref{sec:secure-nip-results}).

We collected a data-set with RAW images from 8 cameras (Table~\ref{tab:cameras}). The photographs come from two public (Raise~\cite{dataset:raise} and MIT-5k~\cite{dataset:fivek}) and from one private data-set. For each camera, we randomly selected 150 images with landscape orientation. These images will later be divided into separate training/validation sets.

\begin{table}[t]
\caption{Digital cameras used in our experiments}
\label{tab:cameras}
\centering
\resizebox{\columnwidth}{!}{
\begin{tabular}{lrrrr}
\toprule 
\textbf{Camera} & \textbf{SR}$^1$ & \#\textbf{Images}$^2$ & \textbf{Source} & \textbf{Bayer}\tabularnewline
\midrule
Canon EOS 5D  & 12 & 864 dng & MIT-5k & RGGB\tabularnewline
Canon EOS 40D & 10 & 313 dng & MIT-5k & RGGB\tabularnewline
Nikon D5100   & 16 & 288 nef & Private & RGGB\tabularnewline
Nikon D700    & 12 & 590 dng & MIT-5k & RGGB\tabularnewline
Nikon D7000   & 16 & $>$1k nef & Raise & RGGB\tabularnewline
Nikon D750    & 24 & 312 nef & Private & RGGB\tabularnewline
Nikon D810    & 36 & 205 nef & Private & RGGB\tabularnewline
Nikon D90     & 12 & $>$1k nef & Raise & GBRG\tabularnewline
\bottomrule
\multicolumn{5}{l}{\footnotesize $^1$ Sensor Resolution in Megapixels [Mpx]} \tabularnewline
\multicolumn{5}{l}{\footnotesize $^2$ RAW file formats: nef (Nikon); dng (generic, Adobe)}
\end{tabular}}
\end{table}

\subsection{Neural Imaging Pipelines}
\label{sec:pipelines}

\begin{figure}
    \centering
    \includegraphics[width=0.85\columnwidth]{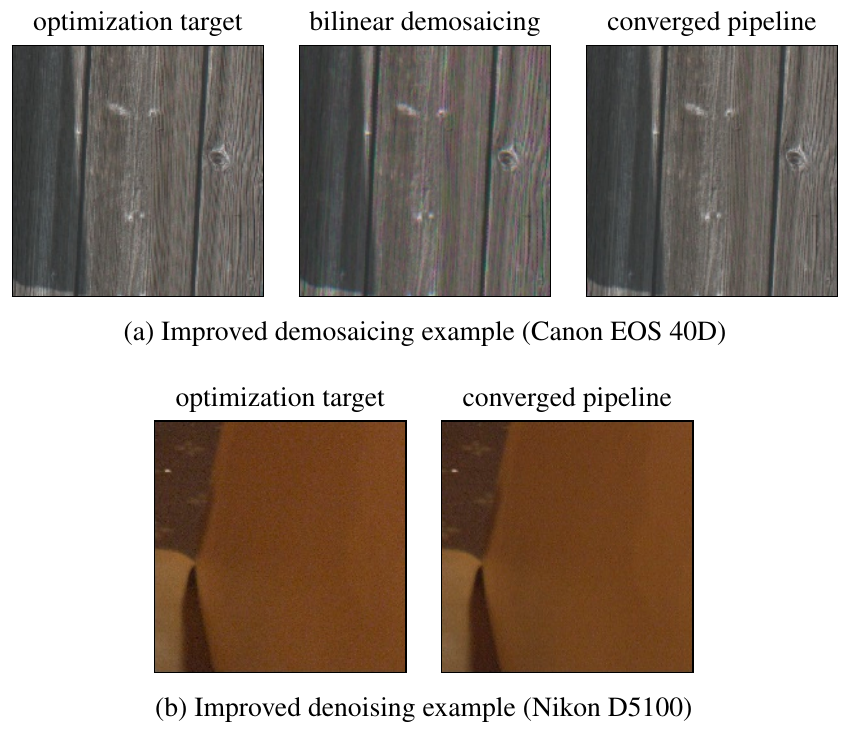}
    \caption{Examples of serendipitous image quality improvements obtained by neural imaging pipelines: (a) better demosaicing performance; (b) better denoising.}
    \label{fig:nip-visual-improvement}
\end{figure}

We considered three NIP models with various complexity and design principles (Table~\ref{tab:pipelines}): \emph{INet} - a simple convolutional network with layers corresponding to successive steps of the standard pipeline; \emph{UNet} - the well-known UNet architecture~\cite{ronneberger2015u} adapted from~\cite{Chen2018}; \emph{DNet} - adaptation of the model originally used for joint demosaicing and denoising~\cite{Gharbi2016}. Details of the networks' architectures are included in the supplement.

We trained a separate model for each camera. For training, we used 120 full-resolution images. In each iteration, we extracted random $128\times128$ patches and formed a batch with 20 examples (one patch per image). For validation, we used a fixed set of $512\times512$~px patches extracted from the remaining 30 images. The models were trained to reproduce the output of a standard imaging pipeline. We used our own implementation based on Python and \emph{rawkit}~\cite{rawkit} wrappers over \emph{libRAW}~\cite{libraw}. Demosaicing was performed using an adaptive algorithm by Menon et al.~\cite{menon2007demosaicing}.

\begin{table}[t]
    \caption{Considered neural imaging pipelines}
    \label{tab:pipelines}
    \centering
    \begin{footnotesize}    
        \begin{tabular}{lccc}
            \toprule 
            & \textbf{INet} & \textbf{UNet} & \textbf{DNet} \tabularnewline
            \midrule
            \textbf{\# Parameters} & 321 & 7,760,268 & 493,976\tabularnewline
            \textbf{PSNR} & 42.8 & 44.3 & 46.2\tabularnewline
            \textbf{SSIM} & 0.989 & 0.990 & 0.995\tabularnewline
            \textbf{Train. speed [it/s]} & 8.80 & 1.75 & 0.66\tabularnewline
            \textbf{Train. time} & 17 - 26 min & 2-4 h & 12 - 22 h\tabularnewline
            \bottomrule
        \end{tabular}
    \end{footnotesize}
    \end{table}

All NIPs successfully reproduced target images with high fidelity. The resulting color photographs are visually indistinguishable from the targets. Objective fidelity measurements for the validation set are collected in Table~\ref{tab:pipelines} (average for all 8 cameras). Interestingly, the trained models often revealed better denoising and demosaicing performance, despite the lack of a denoising step in the simulated pipeline, and the lack of explicit optimization objectives (see Fig.~\ref{fig:nip-visual-improvement}). 

Of all of the considered models, \emph{INet} was the easiest to train - not only due to its simplicity, but also because it could be initialized with meaningful parameters that already produced valid results and only needed fine-tuning. We initialized the demosaicing filters with bilinear interpolation, color space conversion with a known multiplication matrix, and gamma correction with a toy model separately trained to reproduce this non-linearity. The \emph{UNet} model was initialized randomly, but improved rapidly thanks to skip connections. The \emph{DNet} model took the longest and for a long time had problems with faithful color rendering. The typical training times are reported in Table~\ref{tab:pipelines}. The measurements were collected on a Nvidia Tesla K80 GPU. The models were trained until the relative change of the average validation loss for the last 5 dropped below $10^{-4}$. The maximum number of epochs was 50,000. For the \emph{DNet} model we adjusted the stopping criterion to $10^{-5}$ since the training progress was slow, and often terminated prematurely with incorrect color rendering.

\subsection{Image Manipulation}
\label{sec:manipulation}

\begin{figure}
    \centering
    \includegraphics[width=1.0\columnwidth]{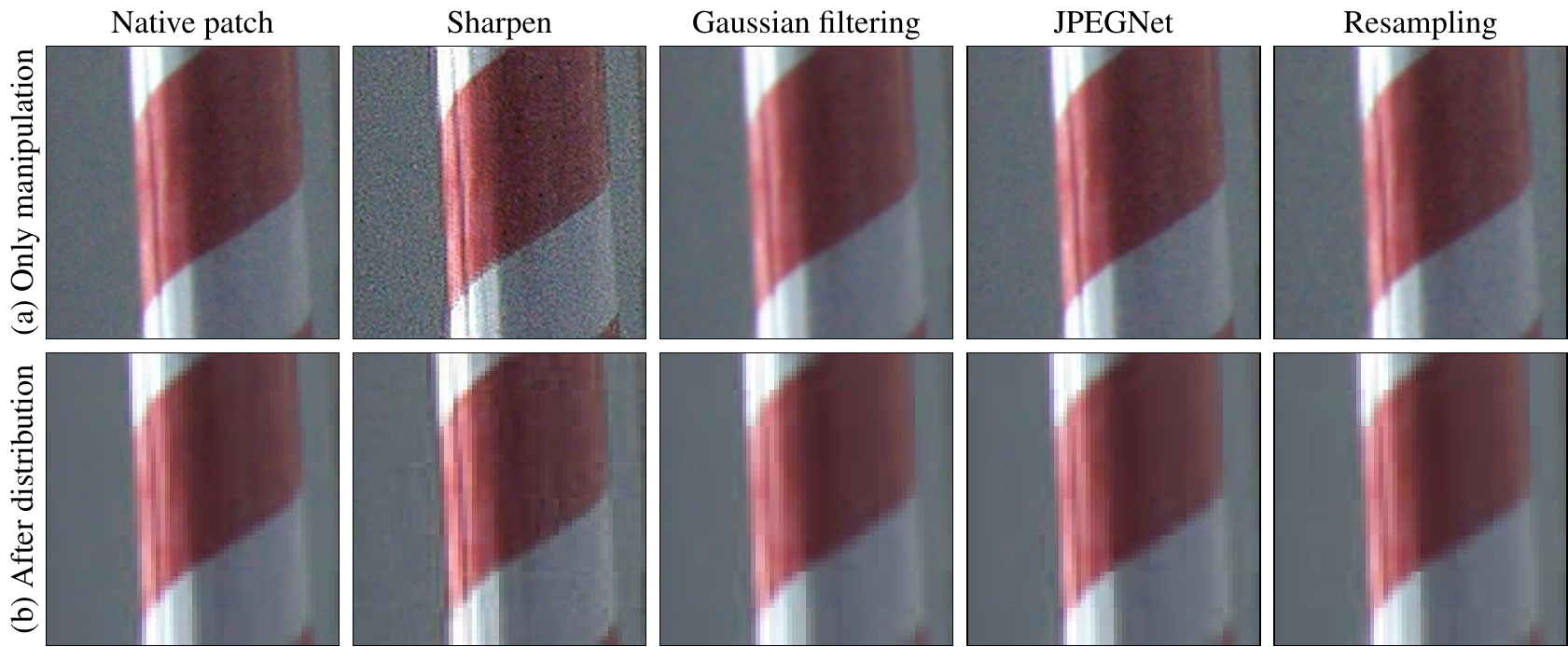}
    \caption{An example image patch with all of the considered manipulations: (a) only manipulation; (b) after the distribution channel (down-sampled and JPEG compressed).}
    \label{fig:post-processing}
\end{figure}

Our experiment mirrors the standard setup for image manipulation detection~\cite{Fan2015,bayar2018constrained,boroumand2018deep}. The FAN simply decides which manipulation class the input patch belongs to. This approximates identification of the last post-processing step, which is often used to mask traces of more invasive modifications. 

We consider four mild post-processing operations: \emph{sharpening} - implemented as an unsharp mask operator with the following kernel: 
    \begin{eqnarray}
        \begin{footnotesize}
        \frac{1}{6}
        \begin{bmatrix}
            -1 & -4 & -1 \\
            -4 & 26 & -4 \\
            -1 & -4 & -1 \\
        \end{bmatrix}
    \end{footnotesize}
    \end{eqnarray}
applied to the luminance channel in the HSV color space; \emph{resampling} - implemented as successive 1:2 down-sampling and 2:1 up-sampling using bilinear interpolation; \emph{Gaussian filtering} - implemented using a convolutional layer with a $5\times5$ filter and standard deviation 0.83; \emph{JPG compression} - implemented using the dJPEG module with sinusoidal rounding approximation and quality level 80. Fig.~\ref{fig:post-processing} shows the post-processed variants of an example image patch: (a) just after manipulation; and (b) after the distribution channel (as seen by the FAN module).

\subsection{FAN Model Validation}
\label{sec:fan-validation}

To validate our FAN model, we initially implemented a simple experiment, where analysis is performed just after image manipulation (no distribution channel distortion, as in~\cite{bayar2018constrained}). We used the \emph{UNet} model to develop smaller image patches to guarantee the same size of the inputs for the FAN model ($128\times128\times3$ RGB images). In such conditions, the model works just as expected, and yields classification accuracy of 99\%~\cite{bayar2018constrained}.

\subsection{Imaging Pipeline Optimization}
\label{sec:secure-nip-results}

\begin{figure}
    \centering
    \includegraphics[width=1.00\columnwidth]{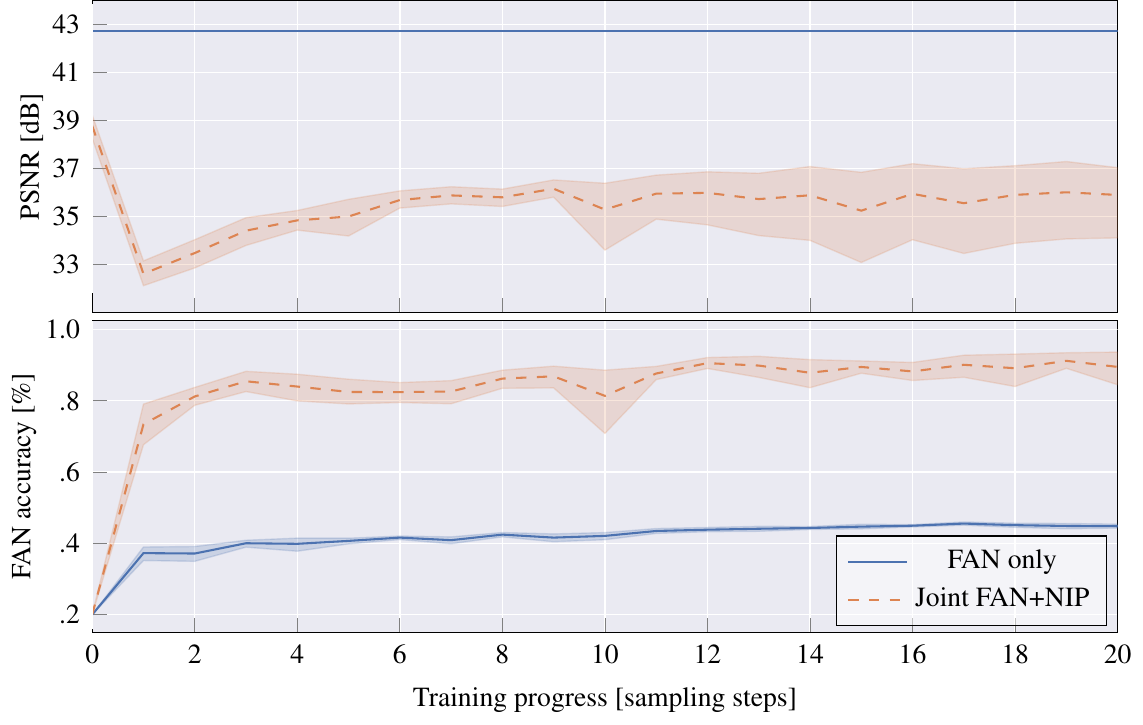}
    \caption{Typical progression of validation metrics (Nikon D90) for standalone FAN training (F) and joint optimization of FAN and NIP models (F+N).}
    \label{fig:training-progress}
\end{figure}

\begin{figure*}
    \centering
    \includegraphics[width=0.95\textwidth]{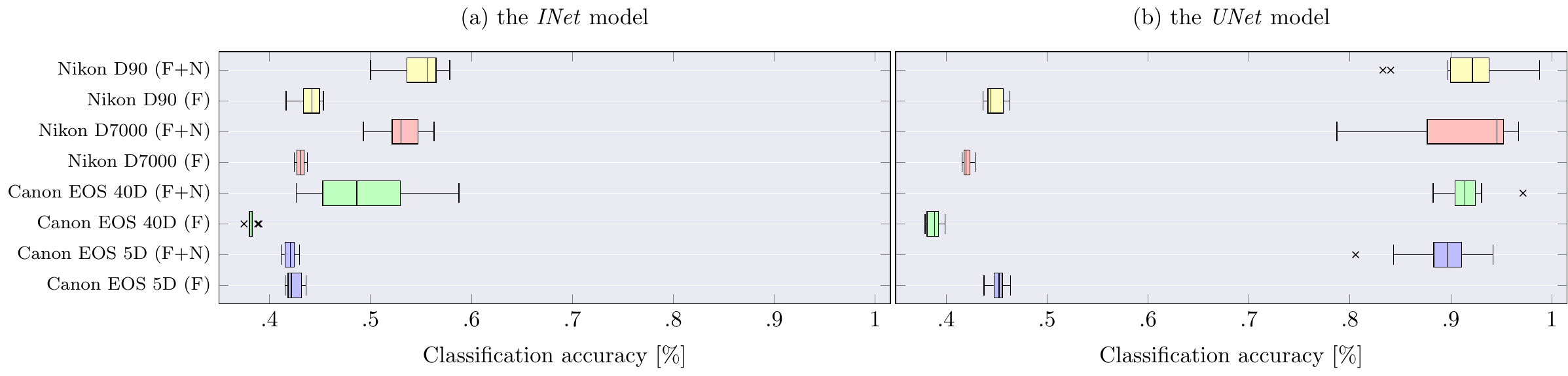}
    \caption{Validation accuracy for image manipulation detection after the distribution channel: (F) denotes standalone FAN training given a fixed NIP; (F+N) denotes joint optimization of the FAN and NIP models.}
    \label{fig:boxplot-accuracy}
\end{figure*}

In this experiment, we perform forensic analysis at the end of the distribution channel. We consider two optimization modes: (F) only the FAN network is optimized given a fixed NIP model; (F+N) both the FAN and NIP models are optimized jointly. In both cases, the NIPs are pre-initialized with previously trained models (Section~\ref{sec:pipelines}). The training was implemented with two separate Adam optimizers, where the first one updates the FAN (and in the F+N mode also the NIP) and the second one updates the NIP based on the image fidelity objective.

Similarly to previous experiments, we used 120 full-resolution images for training, and the remaining 30 images for validation. From training images, in each iteration we randomly extract new patches. The validation set is fixed at the beginning and includes 100 random patches per each image (3,000 patches in total) for classification accuracy assessment. To speed-up image fidelity evaluation, we used 2 patches per image (60 patches in total). In order to prevent over-representation of empty patches, we bias the selection by outward rejection of patches with pixel variance $<$ 0.01, and by 50\% chance of keeping patches with variance $<$ 0.02. More diverse patches are always accepted. 

Due to computational constraints, we performed the experiment for 4 cameras (Canon EOS 40D and EOS 5D, and Nikon D7000, and D90, see Table~\ref{tab:cameras}) and for the \emph{INet} and \emph{UNet} models only. (Based on preliminary experiments, we excluded the \emph{DNet} model which rapidly lost and could not regain image representation fidelity.) We ran the optimization for 1,000 epochs, starting with a learning rate of $10^{-4}$ and systematically decreasing by 15\% every 100 epochs. Each run was repeated 10 times. The typical progression of validation metrics for the \emph{UNet} model (classification accuracy and distortion PSNR) is shown in Fig.~\ref{fig:training-progress} (sampled every 50 epochs). The distribution channel significantly disrupts forensic analysis, and the classification accuracy drops to $\approx 45\%$ for standalone FAN optimization (F). In particular, the FAN struggles to identify three low-pass filtering operations (Gaussain filtering, JPEG compression, and re-sampling; see confusion matrix in Tab.~\ref{tab:confusion}a), which bear strong visual resemblance at the end of the distribution channel (Fig.~\ref{fig:post-processing}b). Optimization of the imaging pipeline significantly increases classification accuracy (over 90\%) and makes the manipulation paths easily distinguishable (Tab.~\ref{tab:confusion}c). Most mistakes involve the \emph{native} and \emph{jpeg} compressed classes. Fig.~\ref{fig:boxplot-accuracy} collects the obtained results for both the \emph{UNet} and \emph{INet} models. While \emph{UNet} delivers consistent and significant benefits, \emph{INet} is much simpler and yields only modest improvements in accuracy - up to $\approx{}55\%$. It also lacked consistency and for one of the tested cameras yielded virtually no benefits. 

\begin{table}[t]
    \caption{Typical confusion matrices (Nikon D90). Entries $\approx{}0$ are not shown; entries $\lessapprox3\%$ are marked with (*).}
    \label{tab:confusion}
    \vspace{4pt}
    \centering
    \resizebox{0.8\columnwidth}{!}{
    \begin{tabular}{lrrrrr}
        \multicolumn{6}{c}{(a) standalone FAN optimization (UNet) $\rightarrow$ 44.2\%} \tabularnewline   
        \diagbox{\textbf{True}}{\textbf{Predicted}} & \rotatebox{90}{\textbf{nat.~}}  &  \rotatebox{90}{\textbf{sha.}}  &  \rotatebox{90}{\textbf{gau.}}  &  \rotatebox{90}{\textbf{jpg}}  &  \rotatebox{90}{\textbf{res.}}  \tabularnewline
        \toprule 
        \textbf{native} & \cellcolor{lime!27} 27& \cellcolor{red!24} 24& \cellcolor{red!18} 18& \cellcolor{red!20} 20& \cellcolor{red!11} 11 \tabularnewline
        \textbf{sharpen} & *& \cellcolor{lime!93} 93& \cellcolor{red!3} 3& *& * \tabularnewline
        \textbf{gaussian} & \cellcolor{red!7} 7& \cellcolor{red!4} 4& \cellcolor{lime!59} 59& \cellcolor{red!12} 12& \cellcolor{red!18} 18 \tabularnewline
        \textbf{jpg} & \cellcolor{red!26} 26& \cellcolor{red!23} 23& \cellcolor{red!18} 18& \cellcolor{lime!21} 21& \cellcolor{red!12} 12 \tabularnewline
        \textbf{resample} & \cellcolor{red!14} 14& \cellcolor{red!7} 7& \cellcolor{red!38} 38& \cellcolor{red!20} 20& \cellcolor{lime!21} 21 \tabularnewline
        \bottomrule
        \tabularnewline

        \multicolumn{6}{c}{(b) joint FAN+NIP optimization (INet) $\rightarrow$ 55.2\% } \tabularnewline        
        \diagbox{\textbf{True}}{\textbf{Predicted}} & \rotatebox{90}{\textbf{nat.~}}  &  \rotatebox{90}{\textbf{sha.}}  &  \rotatebox{90}{\textbf{gau.}}  &  \rotatebox{90}{\textbf{jpg}}  &  \rotatebox{90}{\textbf{res.}}  \tabularnewline
        \toprule 
        \textbf{native} & \cellcolor{lime!41} 41& \cellcolor{red!16} 16& \cellcolor{red!22} 22& \cellcolor{red!18} 18& \cellcolor{red!4} 4 \tabularnewline
        \textbf{sharpen} & \cellcolor{red!7} 7& \cellcolor{lime!85} 85& \cellcolor{red!3} 3& \cellcolor{red!4} 4& * \tabularnewline
        \textbf{gaussian} & \cellcolor{red!18} 18& \cellcolor{red!8} 8& \cellcolor{lime!54} 54& \cellcolor{red!10} 10& \cellcolor{red!10} 10 \tabularnewline
        \textbf{jpg} & \cellcolor{red!41} 41& \cellcolor{red!16} 16& \cellcolor{red!21} 21& \cellcolor{lime!19} 19& \cellcolor{red!4} 4 \tabularnewline
        \textbf{resample} & \cellcolor{red!5} 5& *& \cellcolor{red!14} 14& *& \cellcolor{lime!77} 77 \tabularnewline
        \bottomrule
        \tabularnewline

        \multicolumn{6}{c}{(c) joint FAN+NIP optimization (UNet) $\rightarrow$ 94.0\%} \tabularnewline
        \diagbox{\textbf{True}}{\textbf{Predicted}} & \rotatebox{90}{\textbf{nat.~}}  &  \rotatebox{90}{\textbf{sha.}}  &  \rotatebox{90}{\textbf{gau.}}  &  \rotatebox{90}{\textbf{jpg}}  &  \rotatebox{90}{\textbf{res.}}  \tabularnewline        
        \toprule 
        \textbf{native} & \cellcolor{lime!90} 90& & *& \cellcolor{red!9} 9&  \tabularnewline
        \textbf{sharpen} & & \cellcolor{lime!100} 100& & &  \tabularnewline
        \textbf{gaussian} & & & \cellcolor{lime!97} 97& *& * \tabularnewline
        \textbf{jpg} & \cellcolor{red!13} 13& & \cellcolor{red!3} 3& \cellcolor{lime!84} 84&  \tabularnewline
        \textbf{resample} & & & *& & \cellcolor{lime!99} 99 \tabularnewline
        \bottomrule       
    \end{tabular}
    }
\end{table}

\begin{figure}
    \centering
    \includegraphics[width=1.00\columnwidth]{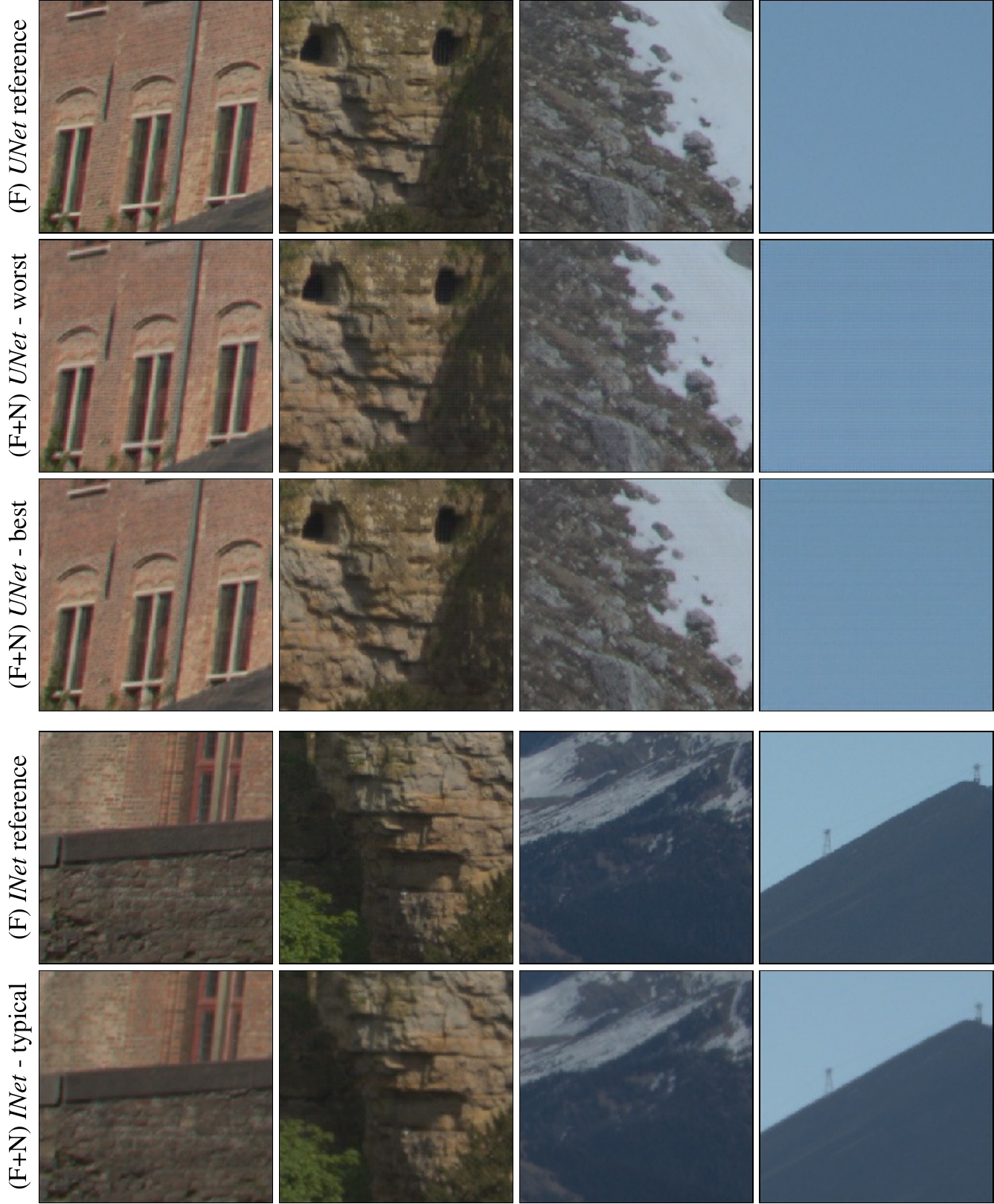}
    \caption{Example image patches developed with joint NIP and FAN optimization (F+N) for Nikon D90. The \emph{UNet} examples show the best/worst artifacts. The \emph{INet} examples show typical outputs.}
    \label{fig:artifacts}
\end{figure}

\begin{table}[t]
    \caption{Fidelity-accuracy trade-off for joint optimization.}
    \label{tab:performance}
    \centering
    \vspace{6pt}
    \resizebox{0.8\columnwidth}{!}{
        \begin{tabular}{lccc}
            \toprule 
            \textbf{Camera} & \textbf{PSNR [dB]} & \textbf{SSIM} & \textbf{Acc. [\%]} \tabularnewline
            \midrule
            & \multicolumn{3}{c}{\textbf{\emph{UNet} model}} \tabularnewline
            \midrule
            \textbf{D90}    & 42.7 $\rightarrow$ 35.9 & 0.990 $\rightarrow$ 0.960 & 0.45 $\rightarrow$ 0.91 \tabularnewline 
            \textbf{D7000}   & 43.0 $\rightarrow$ 35.6 & 0.990 $\rightarrow$ 0.955 & 0.42 $\rightarrow$ 0.91 \tabularnewline
            \textbf{EOS 40D}  & 42.8 $\rightarrow$ 36.1 & 0.990 $\rightarrow$ 0.962 & 0.39 $\rightarrow$ 0.92 \tabularnewline
            \textbf{EOS 5D} & 43.0 $\rightarrow$ 36.2 & 0.989 $\rightarrow$ 0.961 & 0.45 $\rightarrow$ 0.89 \tabularnewline
            \midrule
            & \multicolumn{3}{c}{\textbf{\emph{INet} model}} \tabularnewline
            \midrule
            \textbf{D90}    & 43.3 $\rightarrow$ 37.2 & 0.992 $\rightarrow$ 0.969 & 0.44 $\rightarrow$ 0.55 \tabularnewline 
            \textbf{D7000}   & 40.6 $\rightarrow$ 35.4 & 0.985 $\rightarrow$ 0.959 & 0.43 $\rightarrow$ 0.53 \tabularnewline
            \textbf{EOS 40D}  & 41.6 $\rightarrow$ 33.1 & 0.985 $\rightarrow$ 0.934 & 0.38 $\rightarrow$ 0.50 \tabularnewline
            \textbf{EOS 5D} & 41.5 $\rightarrow$ 40.7 & 0.986 $\rightarrow$ 0.984 & 0.42 $\rightarrow$ 0.42 \tabularnewline
            \bottomrule
        \end{tabular}
    }
\end{table}

The observed improvement in forensic accuracy comes at the cost of image distortion, and leads to artifacts in the developed photographs. In our experiments, photo development fidelity dropped to $\approx$36~dB (PSNR) / 0.96 (SSIM). Detailed results are collected in Tab.~\ref{tab:performance}. The severity of the distortions varied with training repetitions. Qualitative illustration of the observed artifacts is shown in Fig.~\ref{fig:artifacts}. The figure shows diverse image patches developed by several NIP variants (different training runs). The artifacts vary in severity from disturbing to imperceptible, and tend to be well masked by image content. \emph{INet}'s artifacts tend to be less perceptible, which compensates for modest improvements in FAN's classification accuracy. 

\section{Discussion and Future Work}

We replaced the photo acquisition pipeline with a CNN, and developed a fully-differentiable model of the entire acquisition and distribution workflow. This allows for joint optimization of photo analysis and acquisition. Our results show that it is possible to optimize the imaging pipeline to facilitate provenance analysis at the end of complex distribution channels. We observed significant improvements in manipulation detection accuracy w.r.t state-of-the-art classical forensics~\cite{bayar2018constrained} (increase from $\approx{}45\%$ to over 90\%). 

Competing optimization objectives lead to imaging artifacts which tend to be well masked in textured areas, but are currently too visible in flat regions. Severity of the artifacts varies between training runs and NIP architectures, and suggests that better configurations should be achievable by learning to control the fidelity-accuracy trade-off. Since final decisions are often taken by pooling predictions for many patches, we believe a good balance should be achievable even with relatively simple models. Further improvements may also possible by exploring different NIP architectures, or explicitly modeling HVS characteristics. Future work should also assess generalization capabilities to other manipulations, and more complex forensic problems. It may also be interesting to optimize other components of the workflow, e.g., the lossy compression in the channel.

\clearpage

\newpage

\balance

\bibliographystyle{ieee}
\bibliography{bib/references.bib}

\clearpage

\end{document}

\maketitle

\section{Source Code}

To facilitate further research in this direction, and enable reproduction of our results, our neural imaging toolbox is available at \url{https://github.com/pkorus/neural-imaging}.

\section{Implementation Details}
\label{sec:architectures}

We implemented the entire acquisition and distribution pipeline in Python~3 and Tensorflow v. 1.11.0. In all experiments, we used the Adam optimizer with default settings. We proceeded in the following stages:
\begin{enumerate}
    \itemsep0pt
    \item Training and validation of the NIPs.
    \item Validation of the FAN for manipulation detection with no distribution channel between post-processing and forensic analysis.
    \item Joint optimization of the FAN \& NIP models with active distribution channel.
\end{enumerate}
\noindent The NIP models trained in stage (1) were then used in (2) and (3) for NIP initialization to speed-up training. 

\subsection{NIP Architectures}

We considered several NIP models with various levels of complexity. The simplest \emph{INet} model was hand-crafted to replicate the standard imaging pipeline. The architecture (Tab.~\ref{tab:inet}) is a simple sequential concatenation of convolutional layers. Each operation was initialized with meaningful values (see Tab.~\ref{tab:inet}) which significantly sped up convergence. (We also experimented with random initialization, but this led to slower training and occasional problems with color fidelity.) The last two layers were initialized with parameters from a simple 2-layer network trained to approximate gamma correction for a scalar input (4 hidden nodes + one output node).

The \emph{UNet} model (Tab.~\ref{tab:unet}) was the most complex of the considered models, but it delivered fairly quick convergence due to skip connections~\cite{ronneberger2015u}. We adapted the implementation from a recent work by Chen et al.~\cite{Chen2018}. The configuration of skip connections is shown in detail in Tab.~\ref{tab:unet}. All convolutional layers produce tensors of the same spatial dimensions, eliminating the need for cropping. 

We also experimented with two recent networks used for joint demosaicing and denoising (\emph{DNet} - Tab.~\ref{tab:dnet})~\cite{Gharbi2016}, and for general-purpose image processing (\emph{CNet})~\cite{Chen2017}. Overall, the results were unsuccessful. The \emph{DNet} model was able to learn a high-fidelity photo development process, but converged very slowly due to colorization and sharpness artifacts. The trained model proved to be rather fragile, and quickly deteriorated during joint NIP and FAN optimization\footnote{In more recent experiments with explicit regularization of the impact of the NIP's $L_2$ loss, we were able to improve \emph{DNet}'s performance. However, the model still remains more fragile and reaching classification accuracy comparable to \emph{UNet} leads to excessive artifacts. In addition to noise-like artifacts, the model looses edge sharpness. With a visually acceptable distortion, the model yielded accuracy between 70-80\%. More detailed results will be available in an extended version of this work.}. We were unable to obtain satisfactory photo development using the \emph{CNet} model - the validation PSNR stopped improving around 25-27~dB. 

\begin{figure*}
    \centering
    \begin{tikzpicture}[font=\footnotesize]
        \node[draw,line width=0.5mm,inner sep=0pt] (image_a) {\includegraphics[width=\columnwidth]{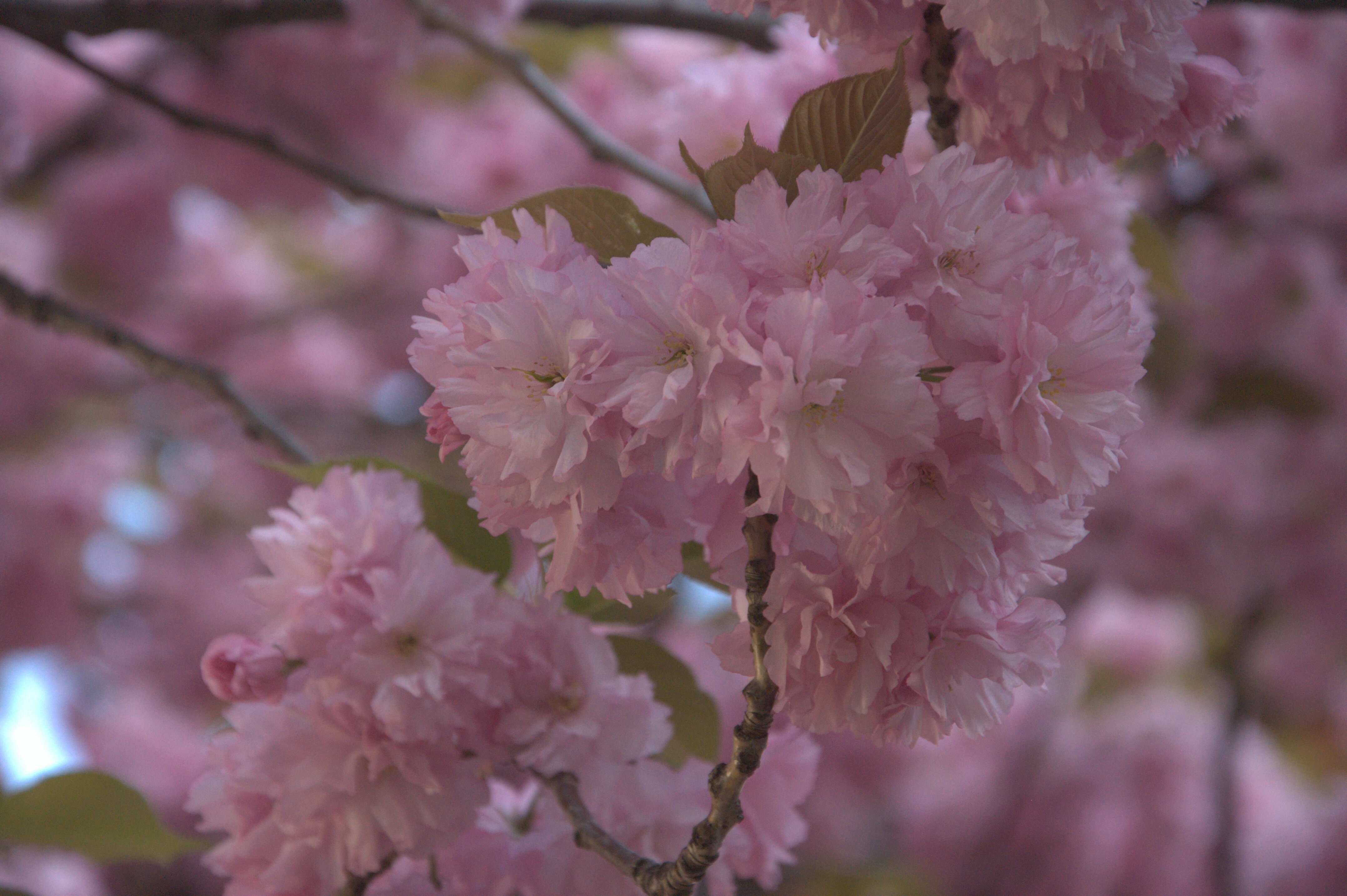}};
        \node[below=0 of image_a] {(a) our implementation of the standard pipeline};
        \node[right=8pt of image_a,draw,line width=0.5mm,inner sep=0pt] (image_b) {\includegraphics[width=\columnwidth]{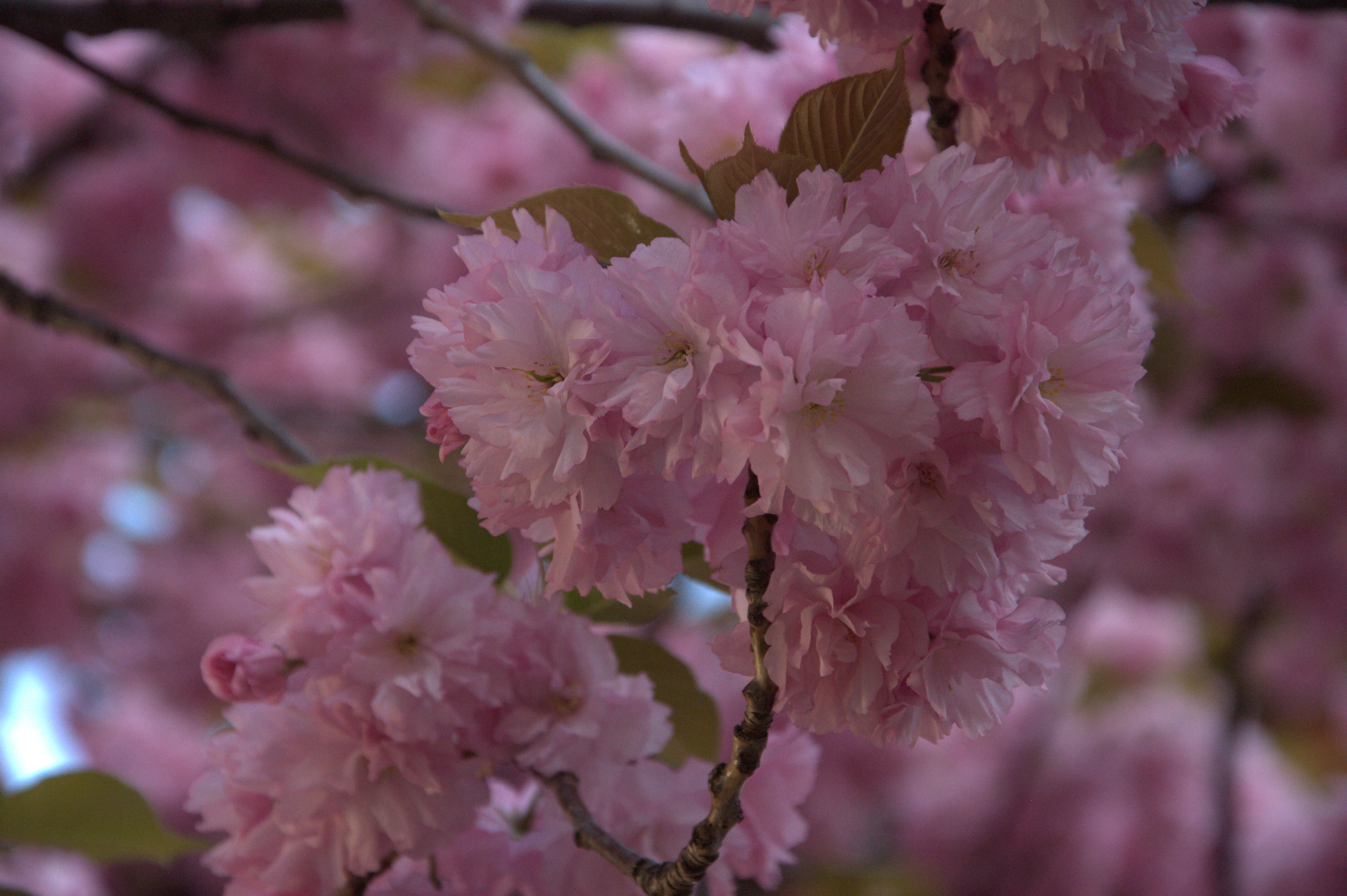}};
        \node[below=0 of image_b] {(b) libRAW with default settings};
        \node[below=16pt of image_a,draw,line width=0.5mm,inner sep=0pt] (image_c) {\includegraphics[width=\columnwidth]{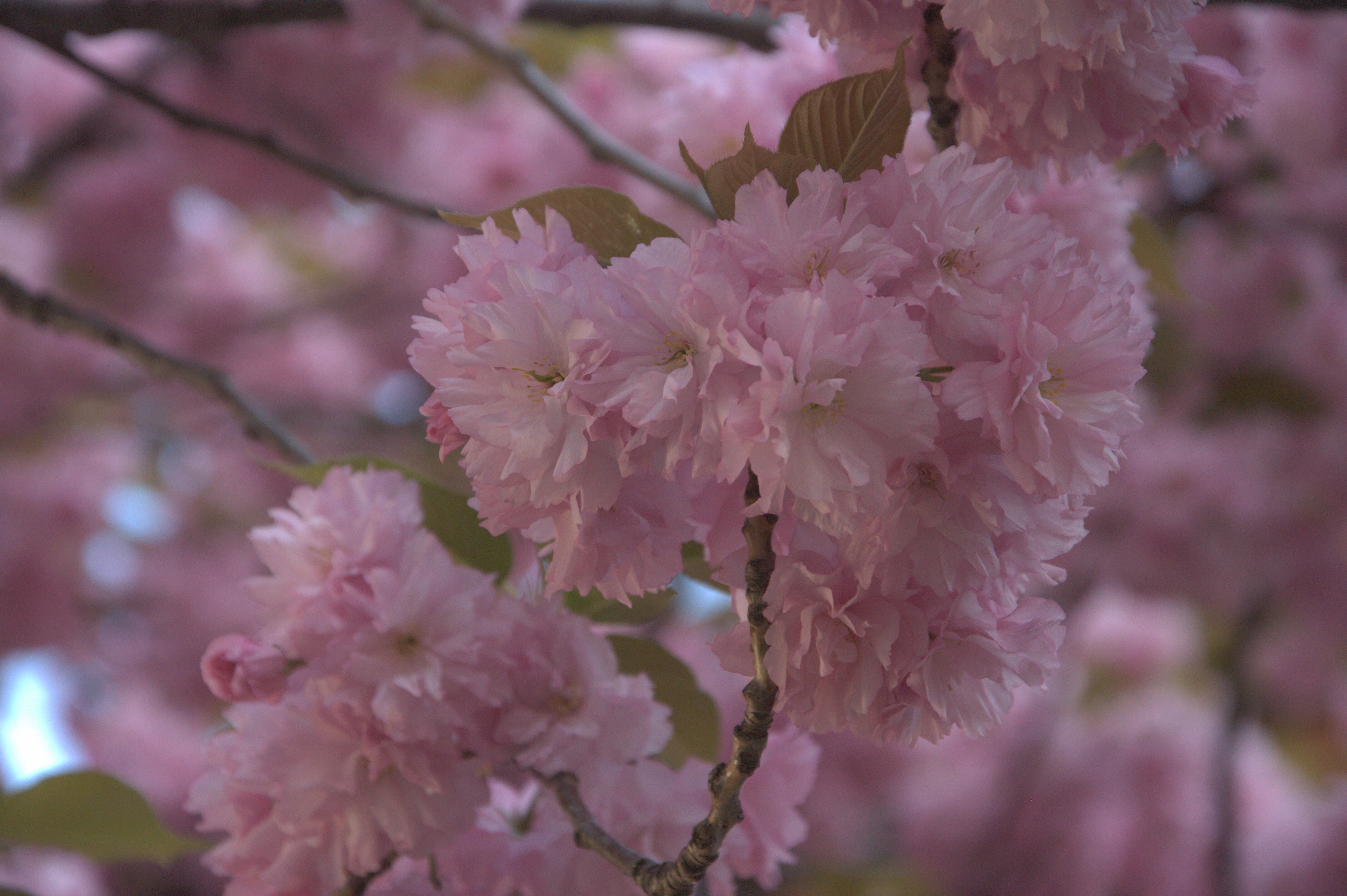}};
        \node[below=0 of image_c] {(c) developed by the INet model};
        \node[below=16pt of image_b,draw,line width=0.5mm,inner sep=0pt] (image_d) {\includegraphics[width=\columnwidth]{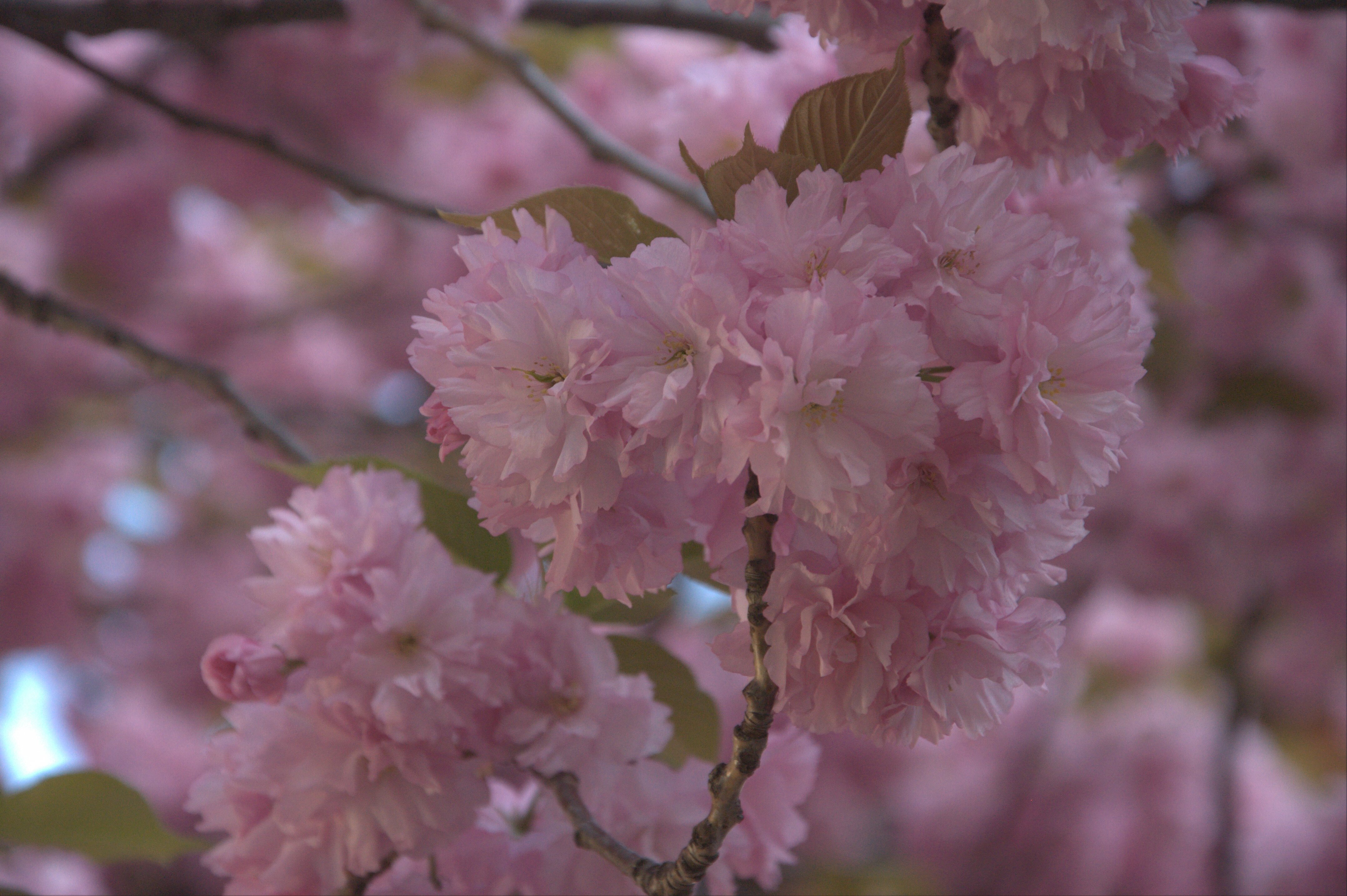}};
        \node[below=0 of image_d] {(d) developed by the UNet model};
        \node[below=16pt of image_c,draw,line width=0.5mm,inner sep=0pt] (image_e) {\includegraphics[width=\columnwidth]{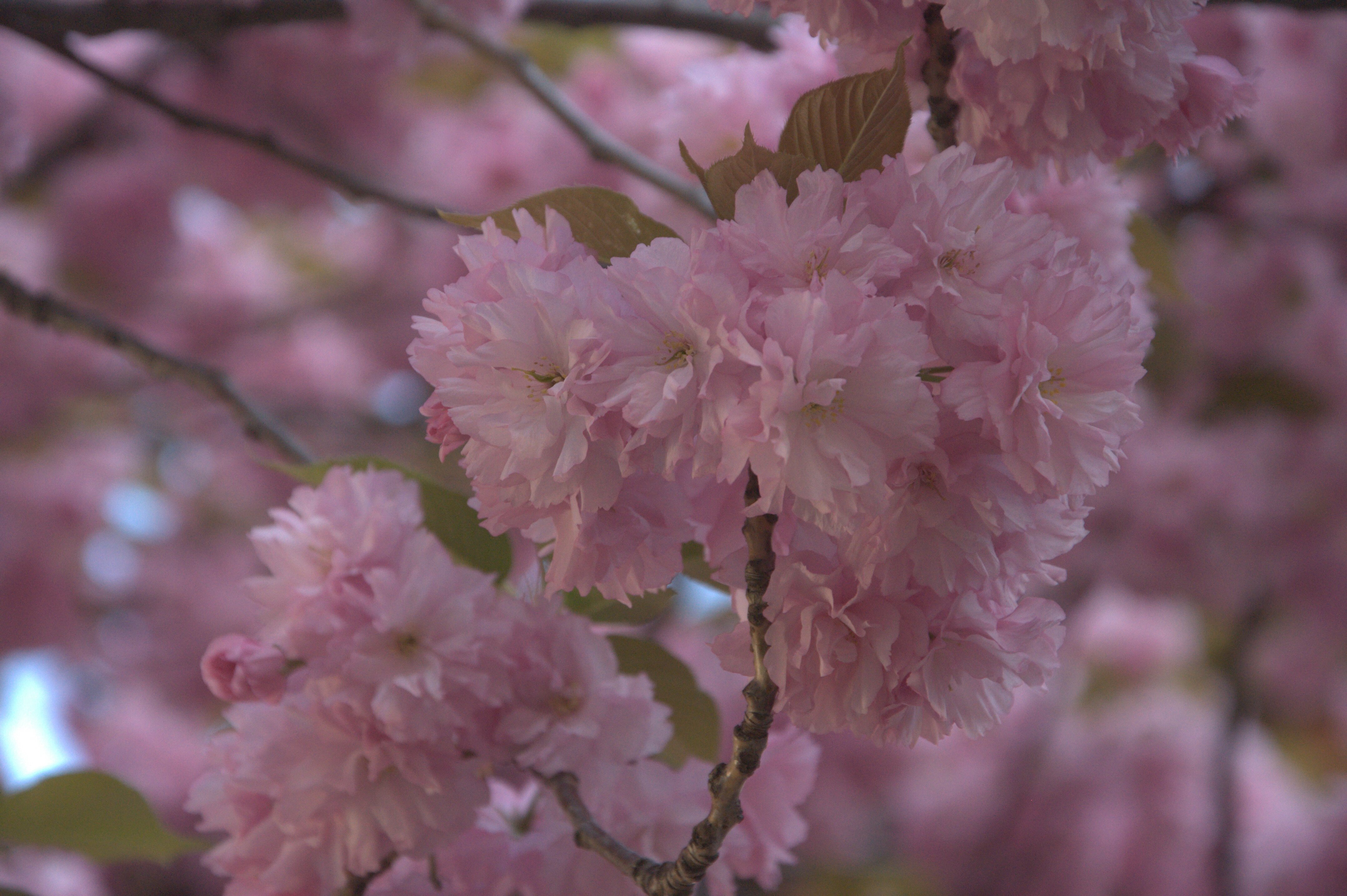}};
        \node[below=0 of image_e] {(e) developed by the DNet model};
        \node[below=0 of image_a.south east,anchor=south east,draw,line width=1mm,white,inner sep=0pt] {\includegraphics[width=0.33\columnwidth]{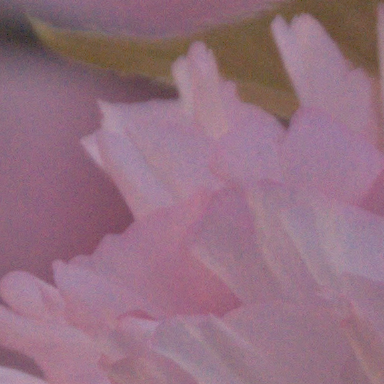}};
        \node[below=0 of image_b.south east,anchor=south east,draw,line width=1mm,white,inner sep=0pt] {\includegraphics[width=0.33\columnwidth]{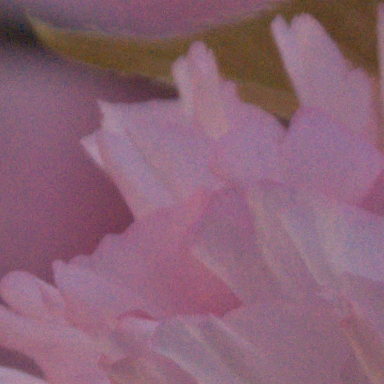}};
        \node[below=0 of image_c.south east,anchor=south east,draw,line width=1mm,white,inner sep=0pt] {\includegraphics[width=0.33\columnwidth]{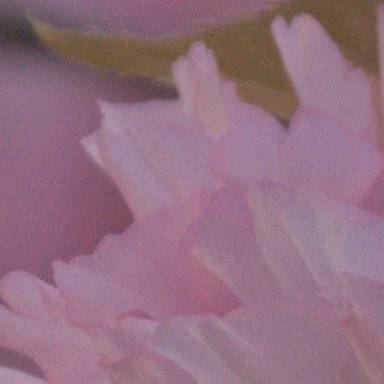}};
        \node[below=0 of image_d.south east,anchor=south east,draw,line width=1mm,white,inner sep=0pt] {\includegraphics[width=0.33\columnwidth]{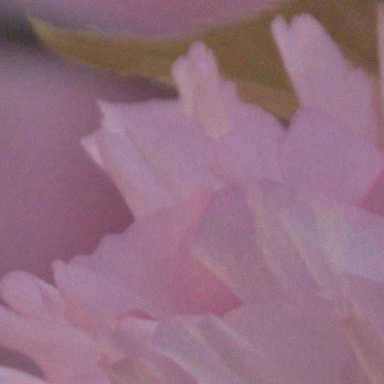}};
        \node[below=0 of image_e.south east,anchor=south east,draw,line width=1mm,white,inner sep=0pt] {\includegraphics[width=0.33\columnwidth]{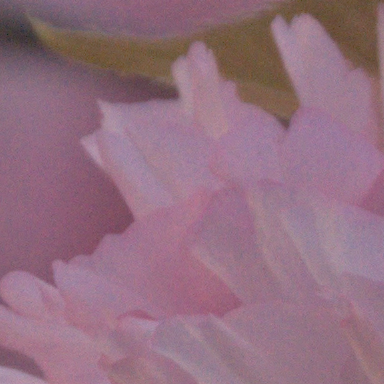}};
    \end{tikzpicture}
    \caption{An example full-resolution (12.3~Mpx) image developed with standard pipelines (ab) and the considered NIPs (cde): image \emph{r23beab04t} from the Nikon D90 camera (Raise dataset~\cite{dataset:raise}). The full-size images are included as JPEGs (quality 85, 4:2:0) to limit PDF size. Close-up patches are included as uncompressed PNG files.}
    \label{fig:full-resolution-example}
\end{figure*}

All models were trained to replicate the output of a manually implemented imaging pipeline. (Detailed, per-camera validation performance measurements are shown in Tab.~\ref{tab:nip-results}.) We used \texttt{rawkit}~\cite{rawkit} wrappers over \texttt{libRAW}~\cite{libraw}. Demosaicing was performed using an adaptive algorithm by Menon et al.~\cite{menon2007demosaicing} from the \texttt{colour-demosaicing} package. We used learning rate of $10^{-4}$ and continued training until the average validation loss for the last 5 epochs changes by more than $10^{-4}$. The maximal number of epochs was set to 50,000. Due to patch-based training, we did not perform any global post-processing (e.g., histogram stretching). In each epoch, we randomly selected patches from full-resolution images, and fed them to the network in batches of 20 examples. We used input examples of size $64 \times 64 \times 4$ (RGGB) which are developed into $128 \times 128 \times 3$ (RGB) color patches. The final networks can work on arbitrary inputs without any changes. Fig.~\ref{fig:full-resolution-example} shows an example full-resolution (12~Mpx) image developed with the standard (ab) and the neural pipelines (cde).

\begin{table}[t]
    \caption{Detailed validation performance statistics for all cameras and all NIPs.}
    \label{tab:nip-results}
    \vspace{4pt}
    \resizebox{1.0\columnwidth}{!}{
    \begin{tabular}{lrrrrrr}
        \toprule
        \textbf{Camera}         & \multicolumn{2}{c}{\textbf{INet}} & \multicolumn{2}{c}{\textbf{UNet}} & \multicolumn{2}{c}{\textbf{DNet}} \tabularnewline
        \cmidrule(lr){2-3} \cmidrule(lr){4-5} \cmidrule(lr){6-7}
        & \textbf{PSNR$^{1}$} & \textbf{SSIM} & \textbf{PSNR$^{1}$} & \textbf{SSIM} & \textbf{PSNR$^{1}$} & \textbf{SSIM} \tabularnewline
        \midrule
        Canon EOS 40D  &  42.7 & 0.987 & 43.6 & 0.990 & 44.5 & 0.992 \tabularnewline
        Canon EOS 5D   &  42.4 & 0.987 & 44.8 & 0.992 & 48.4 & 0.997 \tabularnewline
        Nikon D5100    &  43.7 & 0.989 & 45.3 & 0.990 & 48.1 & 0.996 \tabularnewline
        Nikon D700     &  44.7 & 0.993 & 45.6 & 0.994 & 47.2 & 0.997 \tabularnewline
        Nikon D7000    &  42.3 & 0.989 & 44.4 & 0.992 & 44.9 & 0.994 \tabularnewline
        Nikon D750     &  42.7 & 0.990 & 44.8 & 0.994 & 45.5 & 0.996 \tabularnewline
        Nikon D810     &  39.6 & 0.984 & 41.9 & 0.991 & 43.6 & 0.995 \tabularnewline
        Nikon D90      &  44.6 & 0.993 & 44.4 & 0.991 & 47.7 & 0.997 \tabularnewline
        \bottomrule
        \multicolumn{7}{l}{$^1$ PSNR values in [dB]}
    \end{tabular}}
\end{table}

\begin{table*}[p]
    \caption{The INet architecture: 321 trainable parameters}
    \label{tab:inet}
    \centering
    \begin{footnotesize}
\centering
\begin{tabular}{lllll}
    \toprule     
    \textbf{Operation} & \textbf{Activation} & \textbf{Initialization} & \textbf{Function} & \textbf{Output size} \tabularnewline
    \midrule     
    Input                      & - & - & RGGB feature maps & $N\times \frac{h}{2}\times \frac{w}{2} \times4$\tabularnewline
    $1 \times 1$ convolution   & - & hand-crafted binary sample selection$^{1}$ & Reorganizes data for up-sampling & $N\times \frac{h}{2}\times \frac{w}{2} \times 12 $\tabularnewline
    Depth to space              & - & - & Up-sampling & $N\times h\times w \times 3 $\tabularnewline
    $5 \times 5$ convolution   & - & zero-padded $3\times{}3$ bilinear kernel & Demosaicing & $N\times h\times w \times 3 $\tabularnewline
    $1 \times 1$ convolution   & - & sample color conversion matrix & Color-space conversion (sRGB) & $N\times h \times w \times 3 $\tabularnewline
    $1 \times 1$ convolution   & tanh & pre-trained model & Gamma correction$^{2}$ & $N\times h \times w \times 12 $\tabularnewline
    $1 \times 1$ convolution   & - & pre-trained model & Gamma correction$^{2}$ & $N\times h \times w \times 3 $\tabularnewline
    Clip to [0,1]              & - & - & output RGB image & $N\times h \times w \times 3 $\tabularnewline
    \bottomrule
    \multicolumn{5}{l}{$^{1}$ we disabled optimization of this filter to speed up convergence} \tabularnewline
    \multicolumn{5}{l}{$^{2}$ adapted from a 2-layer network trained separately to approximate gamma correction} \tabularnewline
    \end{tabular}    
\end{footnotesize}    

\end{table*}

\begin{table*}[p]
    \caption{The UNet architecture: 7,760,268 trainable parameters}
    \label{tab:unet}
    \centering
    \begin{footnotesize}
\centering
\begin{tabular}{lllll}
    \toprule     
    \textbf{Operation} & \textbf{Activation} & \textbf{Input} & \textbf{Output} & \textbf{Output size} \tabularnewline
    \midrule     
    Input                      & - & - & $x$ & $N\times h/2 \times w/2 \times4 $\tabularnewline
    $3 \times 3$ convolution   & leaky ReLU & $x$ & $c_{1,1}$ & $N\times h/2 \times w/2 \times 32 $ \tabularnewline
    $3 \times 3$ convolution   & leaky ReLU & $c_{1,1}$ & $c_{1,2}$ & $N\times h/2 \times w/2 \times 32 $ \tabularnewline
    $2 \times 2$ max pooling   & - & $c_{1,2}$ & $p_1$ & $N\times h/4 \times w/4 \times 32 $ \tabularnewline
    \midrule
    $3 \times 3$ convolution   & leaky ReLU & $p_1$ & $c_{2,1}$ & $N\times h/4 \times w/4 \times 64 $ \tabularnewline
    $3 \times 3$ convolution   & leaky ReLU & $c_{2,1}$ & $c_{2,2}$ & $N\times h/4 \times w/4 \times 64 $ \tabularnewline
    $2 \times 2$ max pooling   & - & $c_{2,2}$ & $p_2$ & $N\times h/8 \times w/8 \times 32 $ \tabularnewline
    \midrule
    $3 \times 3$ convolution   & leaky ReLU & $p_2$ & $c_{3,1}$ & $N\times h/8 \times w/8 \times 128 $ \tabularnewline
    $3 \times 3$ convolution   & leaky ReLU & $c_{3,1}$ & $c_{3,2}$ & $N\times h/8 \times w/8 \times 128 $ \tabularnewline
    $2 \times 2$ max pooling   & - & $c_{3,2}$ & $p_3$ & $N\times h/16 \times w/16 \times 128 $ \tabularnewline
    \midrule
    $3 \times 3$ convolution   & leaky ReLU & $p_3$ & $c_{4,1}$ & $N\times h/16 \times w/16 \times 256 $ \tabularnewline
    $3 \times 3$ convolution   & leaky ReLU & $c_{4,1}$ & $c_{4,2}$ & $N\times h/16 \times w/16 \times 256 $ \tabularnewline
    $2 \times 2$ max pooling   & - & $c_{4,2}$ & $p_4$ & $N\times h/32 \times w/32 \times 256 $ \tabularnewline
    \midrule
    $3 \times 3$ convolution   & leaky ReLU & $p_4$ & $c_{5,1}$ & $N\times h/32 \times w/32 \times 512 $ \tabularnewline
    $3 \times 3$ convolution   & leaky ReLU & $c_{5,1}$ & $c_{5,2}$ & $N\times h/32 \times w/32 \times 512 $ \tabularnewline
    $2 \times 2$ strided convolution & - & $c_{5,2}$ & $s_{5}$ & $N\times h/16 \times w/16 \times 256$ \tabularnewline
    \midrule
    $3 \times 3$ convolution   & leaky ReLU & $s_5~|~c_{4,2}$ & $c_{6,1}$ & $N\times h/16 \times w/16 \times 256 $ \tabularnewline
    $3 \times 3$ convolution   & leaky ReLU & $c_{6,1}$ & $c_{6,2}$ & $N\times h/16 \times w/16 \times 256  $ \tabularnewline
    $2 \times 2$ strided convolution & - & $c_{6,2}$ & $s_{6}$ & $N\times h/8 \times w/8 \times 128$ \tabularnewline
    \midrule
    $3 \times 3$ convolution   & leaky ReLU & $s_6~|~c_{3,2}$ & $c_{7,1}$ & $N\times h/8 \times w/8 \times 128 $\tabularnewline
    $3 \times 3$ convolution   & leaky ReLU & $c_{7,1}$ & $c_{7,2}$ & $N\times h/8 \times w/8 \times 128 $\tabularnewline
    $2 \times 2$ strided convolution & - & $c_{7,2}$ & $s_{7}$ & $N\times h/4 \times w/4 \times 64$ \tabularnewline
    \midrule
    $3 \times 3$ convolution   & leaky ReLU & $s_7~|~c_{2,2}$ & $c_{8,1}$ & $N\times h/4 \times w/4 \times 64 $\tabularnewline
    $3 \times 3$ convolution   & leaky ReLU & $c_{8,1}$ & $c_{8,2}$ & $N\times h/4 \times w/4 \times 64 $\tabularnewline
    $2 \times 2$ strided convolution & - & $c_{7,2}$ & $s_{8}$ & $N\times h/2 \times w/2 \times 32$ \tabularnewline
    \midrule
    $3 \times 3$ convolution   & leaky ReLU & $s_8~|~c_{1,2}$ & $c_{9,1}$ & $N\times h/2 \times w/2 \times 32 $\tabularnewline
    $3 \times 3$ convolution   & leaky ReLU & $c_{9,1}$ & $c_{9,2}$ & $N\times h/2 \times w/2 \times 32 $\tabularnewline
    $1 \times 1$ convolution   & - & $c_{9,2}$ & $c_{10}$ & $N\times h/2 \times w/2 \times 12 $\tabularnewline
    \midrule
    Depth to space   & - & $c_{10}$ & $y_{\text{rgb}}$ & $N\times h \times w \times 3 $\tabularnewline
    Clip to [0,1]    & - & $y_{\text{rgb}}$ & $y$ & $N\times h \times w \times 3 $\tabularnewline
    \bottomrule
    \multicolumn{4}{l}{All leaky ReLUs have $\alpha=0.2$} \tabularnewline
    \multicolumn{4}{l}{$|$ denotes concatenation along the feature dimension} \tabularnewline
    \end{tabular}    
\end{footnotesize}    

\end{table*}

\begin{table*}[p]
    \caption{The DNet architecture: 493,976 trainable parameters}
    \label{tab:dnet}
    \centering
    \begin{footnotesize}
\centering
\begin{tabular}{lllll}
    \toprule     
    \textbf{Operation} & \textbf{Activation} & \textbf{Input} & \textbf{Output} & \textbf{Output size} \tabularnewline
    \midrule     
    Input                      & - & - & $c_0$ & $N\times h/2 \times w/2 \times4 $\tabularnewline
    \midrule
    \multicolumn{4}{l}{Repeat for $i$ = 1, 2, \ldots, 14 \{} \tabularnewline
    ~~~~$3 \times 3$ convolution + BN  & ReLU & $c_{i-1}$ & $\hat{c}_{i}$  & $N\times h/2 - 2 \times w/2 - 2 \times 64 $ \tabularnewline
    ~~~~Padding (reflection)       & - & $\hat{c}_{i}$   & $c_{i}$  & $N\times h/2 \times w/2 \times 64 $ \tabularnewline
    \multicolumn{4}{l}{\}} \tabularnewline
    \midrule
    $3 \times 3$ convolution + BN   & ReLU & $c_{14}$  & $\hat{c}_{15}$  & $N\times h/2 - 2 \times w/2 - 2 \times 12 $ \tabularnewline
    Padding (reflection)       & - & $\hat{c}_{15}$  & $c_{15}$  & $N\times h/2 \times w/2 \times 12 $ \tabularnewline
    \midrule
    Depth to space             & - & $c_{15}$   & $f_\text{conv}$    & $N\times h \times w \times 3 $ \tabularnewline
    \midrule
    $1 \times 1$ convolution   & - & $c_{0}$   & $c_{16}$    & $N\times h/2 \times w/2 \times 12 $ \tabularnewline
    Depth to space             & - & $c_{16}$  & $f_\text{bayer}$    & $N\times h \times w \times 3 $ \tabularnewline
    \midrule
    $3 \times 3$ convolution   & ReLU & $f_\text{conv}~|~f_\text{bayer}$ & $\hat{c}_{17}$ & $N\times h - 2 \times w - 2 \times 64 $ \tabularnewline
    Padding (reflection)       & - & $\hat{c}_{17}$  & $c_{17}$  & $N\times h \times w \times 64 $ \tabularnewline
    $1 \times 1$ convolution   & - & $c_{17}$ & $y_{\text{rgb}}$ & $N\times h \times w \times 3 $ \tabularnewline
    Clip to [0,1]              & - & $y_{\text{rgb}}$ & $y$ & $N\times h \times w \times 3 $\tabularnewline
    \bottomrule
    \multicolumn{4}{l}{$|$ denotes concatenation along the feature dimension} \tabularnewline
    \end{tabular}    
\end{footnotesize}    

\end{table*}

\subsection{JPEG Codec Approximation}

The architecture of the dJPEG model is shown in Tab.~\ref{tab:jpegnet}. The network was hand-crafted to replicate the operation of the standard JPEG codec with no chrominance sub-sampling (the 4:4:4 mode). We used standard quantization matrices from the IJG codec~\cite{ijg}. See the main body of the paper for approximation details. The input to the network is an image batch, and should be normalized to [0, 1].

\subsection{The FAN Architecture}

The FAN architecture (Tab.~\ref{tab:fan}) is a fairly standard CNN model with an additional constrained convolution layer recommended for forensics applications~\cite{bayar2018constrained}. In contrast to the study by Bayar and Stamm, who used only the green color channel, we take all color channels (RGB). We also used larger patches for better statistics - in all experiments, the input size is $128 \times 128$~px. 

The network starts with a convolution layer constrained to learn a residual filter. We initialized the layer with the following residual filter (padded with zeros to $5\times{}5$):
\begin{equation}
    \begin{bmatrix}
        -1, -2, -1 \\ 
        -2, 12, -2 \\ 
        -1, -2, -1 \\ 
    \end{bmatrix}.
\end{equation}
We used leaky ReLUs instead of tanh for layer activation, and dispensed with batch normalization due to small network size and fast convergence without it. 

\subsection{Training Details}

In our implementation, we use two Adam optimizers (with default settings) for: (1) updating the FAN (and in joint training also the NIP) based on the cross-entropy loss; (2) updating the NIP based on the image fidelity loss ($L_2$). The optimization steps are run in that order. To ensure comparable loss values, the $L_2$ loss was computed based on images normalized to the standard range [0,255]. Analogously to standalone NIP training, we feed raw image patches extracted from full-resolution images. In each epoch, the patches are chosen randomly, and fed in batches of 20. We start with learning rate of $10^{-4}$ and decrease it by 15\% every 50 epochs. 

\begin{table*}
    \caption{The dJPEG architecture for JPEG codec approximation}
    \label{tab:jpegnet}
    \centering
    \begin{footnotesize}
\centering
\begin{tabular}{llc}
    \toprule 
    \textbf{Operation} & \textbf{JPEG Function} & \textbf{Output size}\tabularnewline
    \midrule
    Input & - & $N\times h\times w\times3$\tabularnewline
    $1\times1$ convolution & RGB $\rightarrow$ YCbCr & $N\times h\times w\times3$\tabularnewline
    Space to depth \& reshapes & Isolate $8\times8$ px blocks & $3N\times8\times8\times B$\tabularnewline
    Transpose \& reshape & - & $3BN\times8\times8$\tabularnewline
    2 $\times$ matrix multiplication & Forward 2D DCT & $3BN\times8\times8$\tabularnewline
    Element-wise matrix division & Divide by quantization matrices & $3BN\times8\times8$\tabularnewline
    Rounding / approximate rounding & Quantization & $3BN\times8\times8$\tabularnewline
    Element-wise matrix multiplication & Multiply by quantization matrices & $3BN\times8\times8$\tabularnewline
    2 $\times$ matrix multiplication & Inverse 2D DCT & $3BN\times8\times8$\tabularnewline
    Transpose \& reshape & - & $3N\times8\times8\times B$\tabularnewline
    Depth to space \& reshapes & Re-assemble $8\times8$ px blocks & $N\times h\times w\times3$\tabularnewline
    $1\times1$ convolution & YCbCr $\rightarrow$ RGB & $N\times h\times w\times3$\tabularnewline
    \bottomrule
    \end{tabular}
\end{footnotesize}    

\end{table*}

\begin{table*}
    \caption{The FAN architecture: 1,341,990 trainable parameters}
    \label{tab:fan}
    \centering
    \begin{footnotesize}
\centering
\begin{tabular}{lllll}
    \toprule     
    \textbf{Operation} & \textbf{Activation} & \textbf{Initialization} & \textbf{Comment} & \textbf{Output size} \tabularnewline
    \midrule     
    Input                      & - & -    & RGB input & $N\times h\times w \times 3$\tabularnewline
    $5 \times 5$ convolution   & - & Standard residual filter$^1$ & Constrained convolution  & $N\times h \times w \times 3 $\tabularnewline
    \midrule
    $5 \times 5$ convolution   & leaky ReLU & MSRA & - & $N\times h \times w \times 32 $\tabularnewline
    $2 \times 2$ max pool      & - & -    & - & $N\times h/2 \times w/2 \times 32 $\tabularnewline
    $5 \times 5$ convolution   & leaky ReLU & MSRA & - & $N\times h/2 \times w/2 \times 64 $\tabularnewline
    $2 \times 2$ max pool      & - & -    & - & $N\times h/4 \times w/4 \times 64 $\tabularnewline
    $5 \times 5$ convolution   & leaky ReLU & MSRA & - & $N\times h/4 \times w/4 \times 128 $\tabularnewline
    $2 \times 2$ max pool      & - & -    & - & $N\times h/8 \times w/8 \times 128 $\tabularnewline
    $5 \times 5$ convolution   & leaky ReLU & MSRA & - & $N\times h/8 \times w/8 \times 256 $\tabularnewline
    $2 \times 2$ max pool      & - & -    & - & $N\times h/16 \times w/16 \times 256 $\tabularnewline
    \midrule
    $1 \times 1$ convolution   & leaky ReLU & MSRA & - & $N\times h/16 \times w/16 \times 256 $\tabularnewline
    \midrule
    global average pooling     & - & -    & - & $N \times 256 $\tabularnewline
    fully connected            & leaky ReLU & MSRA & - & $N\times 512 $\tabularnewline
    fully connected            & leaky ReLU & MSRA & - & $N\times 128 $\tabularnewline
    fully connected            & Softmax & MSRA & Class probabilities & $N\times 5 $\tabularnewline
    \bottomrule
    \end{tabular}    
\end{footnotesize}    

\end{table*}

\bibliographystyle{ieee}
\bibliography{supplement.bib}